\documentclass[10pt,journal,compsoc]{IEEEtran}
\usepackage{times}
\usepackage{epsfig}
\usepackage{graphicx}
\usepackage{amsmath}
\usepackage{amssymb}
\usepackage{algorithm}
\usepackage{algorithmic}
\usepackage{array}
\usepackage{booktabs}
\usepackage{makecell}
\usepackage{float}
\usepackage{diagbox}
\usepackage{changepage}
\usepackage{multirow}
\usepackage[breaklinks=true,bookmarks=false]{hyperref}

\ifCLASSOPTIONcompsoc
  \usepackage[nocompress]{cite}
\else
  \usepackage{cite}
\fi
\ifCLASSINFOpdf
\else
\fi
\begin{document}
%
\title{Video Rain/Snow Removal by Transformed \\ Online Multiscale Convolutional Sparse Coding}

\author{Minghan Li,
        Xiangyong Cao,
        Qian Zhao,
        Lei Zhang,~\IEEEmembership{Fellow,~IEEE,}
        Chenqiang Gao,
        \\Deyu Meng,~\IEEEmembership{Member,~IEEE}
\thanks{ Minghan Li, Xiangyong Cao, Qian Zhao and Deyu Meng (corresponding author) are with the School of Mathematics and Statistics and Ministry of Education Key Lab of Intelligent Networks and Network Security, Xi'an Jiaotong University, Shaanxi, P.R. China.
E-mail: \{liminghan0330, caoxiangyong45\}@gmail.com, \{timmy.zhaoqian, dymeng\}@xjtu.edu.cn.}
\thanks{ Lei Zhang is with the Department of Computing, The Hong Kong Polytechnic University, Hung Hom, Hong Kong.
E-mail: cslzhang@comp.polyu.edu.hk.}
\thanks{ Chenqiang Gao is with the Chongqing Key Laboratory of Signal and Information Processing, Chongqing University of Posts and Telecommunications, Chongqing 400065, China.
E-mail: gaochenqiang@gmail.com.}
}

\IEEEtitleabstractindextext{%
\begin{abstract}
Video rain/snow removal from surveillance videos is an important task in the computer vision community since rain/snow existed in videos can severely degenerate the performance of many surveillance system. Various methods have been investigated extensively, but most only consider consistent rain/snow under stable background scenes. Rain/snow captured from practical surveillance camera, however, is always highly dynamic in time with the background scene transformed occasionally. To this issue, this paper proposes a novel rain/snow removal approach, which fully considers dynamic statistics of both rain/snow and background scenes taken from a video sequence. Specifically, the rain/snow is encoded as an online multi-scale convolutional sparse coding (OMS-CSC) model, which not only finely delivers the sparse scattering and multi-scale shapes of real rain/snow, but also well encodes their temporally dynamic configurations by real-time ameliorated parameters in the model. Furthermore, a transformation operator imposed on the background scenes is further embedded into the proposed model, which finely conveys the dynamic background transformations, such as rotations, scalings and distortions, inevitably existed in a real video sequence. The approach so constructed can naturally better adapt to the dynamic rain/snow as well as background changes, and also suitable to deal with the streaming video attributed its online learning mode. The proposed model is formulated in a concise maximum a posterior (MAP) framework and is readily solved by the ADMM algorithm. Compared with the state-of-the-art online and offline video rain/snow removal methods, the proposed method achieves better performance on synthetic and real videos datasets both visually and quantitatively. Specifically, our method can be implemented in relatively high efficiency, showing its potential to real-time video rain/snow removal.
\end{abstract}

}

\maketitle

\IEEEdisplaynontitleabstractindextext

%
\IEEEpeerreviewmaketitle

\section{Introduction}\label{sec:introduction}
\IEEEPARstart{V}{ideos} captured from outdoor surveillance system are often contaminated by rain or snow, which has a negative effect on the perceptual quality and tends to degrade the performance of subsequent video processing tasks, such as human detection~\cite{Dalal05Histograms}, person re-identification~\cite{Farenzena07PersonRe-ID}, object tracking~\cite{mukhopadhyay2014combating} and scene analysis~\cite{Itti98Scenanalysis}. Thus, removing rain and snow from surveillance videos is an important video pre-processing step and has attracted much attention in the computer vision community.

In recent decades, various methods have been proposed for removing rain from a video. The earliest video rain removal approach was proposed based on the photometry property of rain~\cite{Garg2004Detection}. After that, more methods taking advantage of the essential physical characteristics of rain, such as photometric appearance~\cite{Garg2005When}, chromatic consistency~\cite{Zhang06Rain}, shape and brightness~\cite{Barnum07} and spatial-temporal configurations~\cite{Tripathi2011A}, were introduced to better separate rain streaks from the background of videos. However, these methods don't utilize the prior knowledge of video structure, such as spatial smoothness of foreground objects and temporal similarity of background scenes, and thus cannot always obtain satisfactory performance especially in complex scenes. In recent years, low-rank models~\cite{Chen13A} show a great potential for this task and always achieve state-of-the-art performance due to their better consideration of video structure prior knowledge both in foreground and background. Specifically, these methods not only use the low-rank structure for the background, but also fully facilitate the prior knowledge of the rain, such as sparsity and spatial smoothness~\cite{ren2017cvpr,wei2017should}. Very recently, deep learning based methods have also been proposed for this task. These methods address the problem of video rain removal by constructing deep recurrent convolutional networks~\cite{Liu18Erase} or deep convolutional network~\cite{Chen18Robust} and implement the task in a popular end-to-end learning manner.

\begin{figure*}[!htbp]\vspace{-2mm}
\centering
\includegraphics[width=0.9\linewidth]{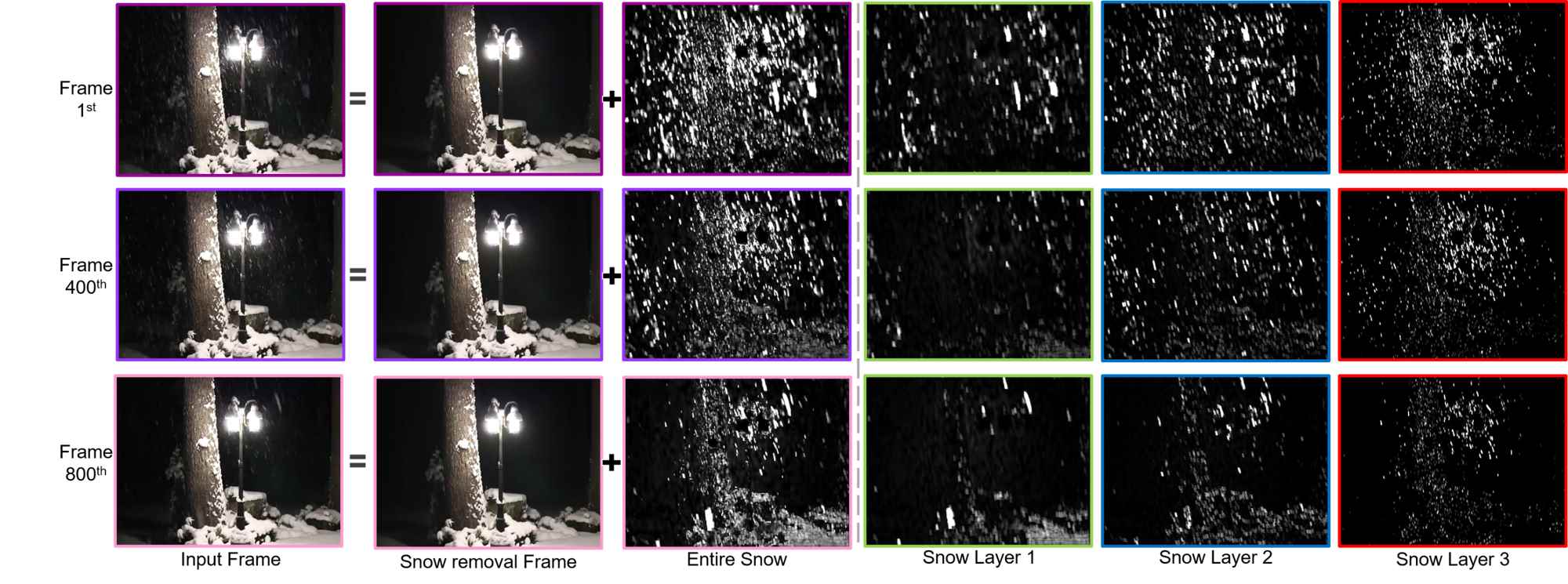}\vspace{-5mm}
\caption{The first column: Three frames in a video with snow gradually varied from heavy to light. The second and third columns: background scenes and snow of the frames obtained by the OTMS-CSC method. The fourth to sixth columns: three snow layers separated by OTMS-CSC.}
\label{fig:online}\vspace{-4mm}
\end{figure*}

Albeit achieving good progress, most of current methods are implemented on a pre-fixed length of videos and assume consistent rain/snow shapes under static background scenes. This, however, is evidently deviated from the real scenarios. On one hand, the rain/snow contained in a video sequence is generally with configurations changed constantly along time, as typically shown in Fig. \ref{fig:online}. On the other hand, the background scene in video is also always dynamic, and inevitably contains timely transformations such as translation, rotation, scaling and distortion, due to camera jitters. Lacking considerations to such dynamic characteristics inclines to degenerate the performance of current methods in such real cases. Besides, as the dramatically increasing surveillance cameras installed all over the world, the real video is always coming online as a streaming format. Most current methods, however, are implemented/trained on a prefixed video sequence, and thus cannot finely and efficiently adapt to such kinds of streaming videos continually and endlessly coming in time. These issues have hampered the availability of existing methods in real applications and thus is worthy to be specifically investigated.

Against the aforementioned issues, this paper proposes a new video rain/snow removal method by fully encoding the dynamic statistics of both rain/snow and background scenes in a video along time into the model, and realizing it with an online mode to make it potentially available to handle constantly coming streaming video sequence. Specifically, inspired by the multi-scale convolutional sparse coding (MS-CSC) model designed for video rain removal (still for static rain) previously proposed in \cite{Li18Multiscale}, which finely delivers the sparse scattering and multi-scale shapes of real rain, this work encodes the dynamic temporal changing tendency of rain/snow as a dynamic MS-CSC framework by timely parameter amelioration for the model in an online implementation manner. Besides, a transformation operator capable of being adaptively updated along time is imposed on the background scenes to finely fit the dynamic background transformations existed in a video sequence. All these knowledge are formulated into a concise maximum a posterior (MAP) framework, which can be easily solved by alternative optimization technique.

In all, the contribution of this work can be mainly summarized as follows: 1) An online MS-CSC model is specifically designed for encoding dynamic rain/snow with temporal variations. The model is formulated as a concise probabilistic framework, where the feature map representing rain/snow knowledge of each video frame is gradually ameliorated under regularization of a penalty for enforcing them close to those calculated from the previous frames. In this manner, the insightful dynamic rain properties, i.e., the correlation and distinctiveness of rain/snow along different video frames, can be finely delivered. 2) An affine transformation operator is further embedded into the proposed model, and can be automatically adjusted to fit a wide range of video background transformations. This makes the method more robust to general camera movements, like rotation, translation, scaling or distortion. 3) The adopted online learning manner makes the method possess fixed space complexity all along but not gradually increasing ones (mostly to infinity) as most conventional methods, and fixed time complexity for any fixed length of newly coming frames. This guarantees the feasibility of our method on any length of video sequence, and  provides potential for the method to handle real streaming data. 4) The superiority of the proposed method in robustness and efficiency are comprehensively substantiated by experiments implemented on synthetic and real videos, including those with evident rain/snow variations and/or dynamic camera jitters, both visually and quantitatively, as compared with other state-of-the-art methods. Specifically, the performance of our method, directly executed on the streaming video sequence, can exceed those deep learning ones, requiring more pre-collected training data sources (pairs of rainy/snowy and corresponding clean video frames). This, to a certain extent, shows that the popularly employed data-driven deep learning methodology, requiring dominant source of supervised training samples and computation powers, might not be the only fashion for solving any computer vision tasks.
It might be still necessary and important for elaborately designing probabilistic models through possibly thoroughly understanding the investigated problem and application scenarios.

The rest of paper is organized as follows. Section 2 introduces the related works. Section 3 reviews the offline MS-CSC model suitable for removing static rain and proposes the online transformed MS-CSC model as well as its solving algorithm. Section 4 demonstrates experimental results on synthetic and real rainy/snowy videos to substantiate the superiority of the proposed method. Finally, conclusions are drawn in Section 5.

\vspace{-1mm}
\section{Related Works}
In this section, we give a brief review on the methods of video rain and snow removal. The related developments on single image rain and snow removal, multi-scale modeling and video alignment are also introduced for literature comprehensiveness.

\vspace{-1mm}
\subsection{Video Rain and Snow Removal Methods}
Garg and Nayar~\cite{Garg2004Detection} made the earliest study on the photometric appearance of rain drops and developed a rain detection method by utilizing a linear space-time correlation model. To better reduce the effects of rain before camera shots in images/videos, Garg and Nayar~\cite{Garg2005When,Garg2007Vision} further proposed a method by adjusting the camera parameters such as field depth and exposure time.

In the past years, more physical intrinsic properties of rain streaks have been explored and formulated in algorithm designing. For example, Zhang et al.~\cite{Zhang06Rain} incorporated both chromatic and temporal properties and utilized K-means clustering for distinguishing background and rain streaks from videos. Later, Barnum et al.~\cite{Barnum07} first considered the impact of snow on videos. They derived a physical model for representing raindrops and snowflakes and used them to determine the general shape and brightness of a single streak. The streak model combined with the statistical properties of rain and snow can then conduct how they affect the spatial-temporal frequencies of an image sequence. To enhance the robustness of rain removal, Barnum et al.~\cite{Barnum10} employed the regular visual effects of rain and snow in global frequency information to approximate rain streaks as a motion-blurred Gaussian. Afterwards, to integrate more prior knowledge of the task, Jiang et al.~\cite{jiang2017cvpr} proposed a tensor-based video rain streak removal approach by considering the sparsity of rain streaks, smoothness along the raindrops and the rain-perpendicular direction, and global and local correlation along time direction.

In recent years, low-rank based models have drawn more research attention for the task of video rain/snow removal. Chen et al.~\cite{Chen13A} first investigated spatial-temporal correlation among local patches with rain streaks and used low-rank term to help extract rain streaks from a video. Later, Kim et al.~\cite{Kim15} proposed a rain and snow removal method which is also designed based on temporal correlation and low-rank matrix completion. This method uses extra supervised knowledge (images/videos with/without rain streaks) to help training a rain classifier. To further exclude false candidates, Santhaseelan et al.~\cite{santhaseelan2015utilizing} used local phase congruency to detect rain and applied chromatic constrain. To deal with heavy rain and snow in dynamic scenes, Ren et al.~\cite{ren2017cvpr} divided rain into sparse and dense ones based on the low-rank hypothesis of the background. Based on the low-rank background assumption, Wei et al.~\cite{wei2017should} further encoded rain streaks as a patch-based mixture of Gaussians. Such stochastic manner for encoding rain streaks could make the method deliver a wider range of rain information.

Very recently, motivated by the booming of deep learning (DL) techniques, several DL methods also appeared for the task. Liu et al.~\cite{Liu18Erase} addressed the problem by constructing deep recurrent convolutional networks, which builds a joint recurrent rain removal and reconstruction network that seamlessly integrates rain degradation classification, spatial texture appearances based rain removal, and temporal coherence based background detail reconstruction. Meanwhile, Chen et al.~\cite{Chen18Robust} proposed a deep derain framework which applies superpixel segmentation to decompose the scene into depth consistent units. Alignment of scene contents are done at the super-pixel level to handle the videos with highly complex and dynamic scenes.

\vspace{-1mm}
\subsection{Single Image Rain and Snow Removal Methods}
For literature comprehensiveness, we also briefly review the rain/snow removal methods for a single image. Kang et al.~\cite{kang2012automatic} firstly formulated the problem as an image decomposition problem based on morphological component analysis, which achieves rain component from the high frequency part of an image by using dictionary learning and sparse coding. Later, Luo et al.~\cite{luo2015removing} built a nonlinear screen blend model based on discriminative sparse codes. Besides, Ding et al.~\cite{ding2016single} designed a guided $L_{0}$ smoothing filter to obtain a coarse rain-free or snow-free image, and Li et al.~\cite{Li2016cvpr} utilized patch-based GMM priors to distinguish and remove rain from background in a single image. Wang et al.~\cite{wang2017hierarchical} designed a 3-layer hierarchical scheme to classify the high-frequency part into rain/snow and non-rain/snow components. Gu et al.~\cite{Gu2017JACS} jointly analyzed sparse representation and synthesis sparse representation to encode background scene and rain streaks. Meanwhile, Zhang et al.~\cite{zhang17convolutional} learned a set of generic sparsity-based and low-rank representation-based convolutional filters for efficiently representing background and rain streaks in an image.

Recently, DL-based methods represent the new trend for this task. Fu et al.~\cite{fu2017clearing} firstly developed a deep CNN model to extract discriminative features of rain in high frequency layer of an image. The training pairs are constructed based on the whole image. Later, Fu et al.~\cite{fu2017removing} constructed the training pairs by using image patches and utilized the res-net as the classifier. Zhang et al.~\cite{Zhang2017idcGan} first proposed a derain network based on generative adversarial network for single image derain. Yang et al.~\cite{Yang2017deep} designed a multi-task DL architecture that learns the binary rain streak map, the appearance of rain streaks and the clean background. Liu et al.~\cite{liu2018desnownet} proposed a multistage and multi-scale network to deal with the removal of translucent and opaque snow particles. Very recently, Yang et al.~\cite{yang2019joint} constructed a contextualized deep network, which incorporates a binary rain map indicating rain-streak regions, and accommodates various shapes, directions, and sizes of overlapping rain streaks as well as rain accumulation to model heavy rain.

Although these image-based methods can also deal with rain/snow removal in a video via a rough frame-by-frame manner, the missing use of the important temporal information for such a specific task inclines to make the video-based methods perform significantly better than image-based ones.

\vspace{-1mm}
\subsection{Multi-scale Approaches}
Since multi-scale represents a general essence of various visual concepts, multi-scale approaches have been applied to wide range of computer vision tasks. E.g., for image segmentation, Baatz et al.~\cite{Baatz2000An} used a scale parameter to control the average image object size, making the method adaptable to different scales of interests. For image quality assessment, Wang et al.~\cite{wang2004image} proposed a multi-scale structural similarity method and developed an image synthesis method to calibrate the parameters that define the relative importance of different scales. To improve the invariance of CNN activations, Gong et al.~\cite{Gong14Multi-scale} presented a simple but effective scheme to design a multi-scale orderless pooling regime. For dense prediction, Yu et al.~\cite{Yu16Multiscale} developed a convolutional network module using dilated convolutions to systematically aggregate multi-scale contextual information without losing resolution.

\vspace{-1mm}
\subsection{Alignment Approaches for Videos}
Since camera jitter tends to damage the low-rank background structure of a video, we always need to align the transformed videos to accurately extract the low-rank background. Many alignment methods have been attempted to this issue. For example, Zhang et al.~\cite{Zhang12TILT}  proposed an approach to directly extract certain 3D invariant structures through their 2D images by undoing the (affine or projective) domain transformations.
Zhang et al.~\cite{Zhang13Simultaneous} further proposed a general method for recovering low-rank 3-order tensors, which introduced auxiliary variables and relaxed the hard equality constraints by the ADMM method. Yong et al.~\cite{Yong2107Robust} proposed an alignment method for aligning the video background based on optimizing a supplemental affine transformation operator, and applied it to the task of dynamic background subtraction.

\vspace{-1mm}
\section{Online Transformed MS-CSC model for Dynamic Video Rain/Snow Removal}
This work is inspired by our previous conference work~\cite{Li18Multiscale}, proposing an offline multi-scale convolutional sparse coding (MS-CSC) model, specifically designed for rain removal issue (with consistent rain temporarily) in a fixed length of video sequence. We thus first introduce the formulation of this offline model.

\vspace{-1mm}
\subsection{Offline MS-CSC Model}
Let $\mathcal{X}\in R^{h\times w\times n}$ denote the input video, where $h, w,$ and $n$ represent the height, width and the number of frames, respectively. We assume that the video $\mathcal{X}$ is decomposed as:
\vspace{-1mm}
\begin{equation}\label{decomp}
\vspace{-1mm}
\mathcal{X=B+F+R+E},
\end{equation}
where $\mathcal{B},\mathcal{F},\mathcal{R},\mathcal{E}\in R^{h\times w\times n}$ represent background scene, rain layer, moving objects, and background noise of the video, respectively. These parts can then be modeled separately as follows \cite{Li18Multiscale}.

\textbf{\textit{Background Modeling}}: For a fixed length of video sequence captured from a surveillance camera, the background tends to keep steady over the frames, and thus can be rationally assumed to be resided on a low-dimensional subspace~\cite{meng2013robust,zhao15l1norm,zhao14robust,cao2016robust}, leading to its low-rank matrix factorization representation as:
\begin{equation}
\mathcal{B}=\mathrm{Fold}(UV^T),
\end{equation}
where $U, V\in R^{n\times r},\ d=hw,\ r< \mathrm{min}(d,n)$. The operation `Fold' refers to fold up each matrix column into the corresponding frame matrix, and thus $\mathcal{B}$ is a tensor with the same size as $\mathcal{X}$.

\textbf{\textit{Rain Layer Modeling}}: Since rain in a video contain repetitive local patterns sparsely scattering over different areas, and also exhibits multi-scale property due
to its occurrence positions with different distances to the
cameras, multi-scale convolutional sparse coding (MS-CSC)~\cite{Zeiler10deconvolutional} is thus utilized to model rain as follows:
\vspace{-1mm}
\begin{equation}
\vspace{-1mm}
\mathcal{R}=\sum\limits_{k=1}^K \sum\limits_{s=1}^{s_k} D_{ks}\otimes \mathcal{M}_{ks},\label{rain}
\end{equation}
where {\small $\mathcal{M}=\{\mathcal{M}_{ks}\}_{k,s=1}^{K,s_k}\subset R^{h\times w\times n}$} is a set of feature maps that approximate
the rain streak positions, and {\small $D=\{D_{ks}\}_{k,s=1}^{K,s_k}\subset R^{p_k\times p_k}$} denotes the filters representing the repetitive local patterns of rain streaks. {\small $K$} and $s_k$ denote the numbers of entire filters and filters at the $k$-th scale, respectively. Considering the sparsity of feature maps, the $L_1$-penalty \cite{meng12improve} is utilized to regularize them.

\textbf{\textit{Moving objects Modeling}}: Motivated by the work~\cite{wei2017should}, Markov random field (MRF) is used to explicitly detect the moving objects. Let $\mathcal{H}\in R^{h\times w\times n}$ be a binary tensor denoting the moving object support:
\vspace{-1mm}
\begin{equation}
\vspace{-1mm}
\mathcal{H}_{ijn}=
\begin{cases}
    1, & \text{location} ~ (i,j,n) ~ \text{is moving objects},\\
    0, & \text{location} ~ (i,j,n) ~ \text{is background}, \\
\end{cases}
\end{equation}
and $\mathcal{H}^\bot$ be the complementary of $\mathcal{H}$ (i.e., $\mathcal{H}+\mathcal{H}^\bot=\mathbf{1}$, $\mathbf{1}$ is a tensor with all elements as 1). Eq.(\ref{decomp}) is then reformulated as:
\begin{equation}
\mathcal{X}=\mathcal{H}^\bot\circ \mathcal{B}+\mathcal{H}\circ \mathcal{F}+\mathcal{R}+\mathcal{E},
\end{equation}
where $\circ$ denotes the element-wise multiplication. Since moving objects always exhibit smooth property, total variation (TV) penalty is adopted to regularize them. Additionally, considering the sparse feature and continuous shapes along both space and time of moving object, $L_1$-penalty and weighted 3-dimensional total variation (3DTV) penalty are both employed to regularize the moving objects support $\mathcal{H}$ simultaneously.

By assuming that the background noise $\mathcal{E}$ follows an i.i.d. Gaussian, we can then integrate the aforementioned three models imposed on background, rain streak and moving objects to get the MS-CSC model for offline video rain removal as follows \cite{Li18Multiscale}:
\begin{eqnarray}\label{mcsc}
\begin{split}
\min\limits_\Theta \mathcal{L}(\Theta)&= \parallel \mathcal{X}-\mathcal{H}^\bot\circ \mathcal{B}-\mathcal{H}\circ \mathcal{F}-R\parallel_F^2 +\lambda\parallel \mathcal{F} \parallel_{TV}\nonumber\\
&+\alpha \parallel \mathcal{H}\parallel_{3DTV}+\beta\parallel \mathcal{H}\parallel_1+b\sum\limits_{k=1}^K \sum\limits_{s=1}^{n_k} \parallel \mathcal{M}_{ks}\parallel_1 \nonumber\\
s.t.\qquad & \mathcal{B} = \mathrm{Fold}(U^TV) \nonumber\\
\qquad &\mathcal{R}=\sum\limits_{k=1}^K\sum\limits_{s=1}^{s_k} D_{ks}\otimes \mathcal{M}_{ks},~~ \parallel D_{ks}\parallel_F^2 \leq 1,
\end{split}
\end{eqnarray}
where $\Theta=\{D,\mathcal{M},\mathcal{H},\mathcal{F},U,V,\mathcal{R}\}$ are the variables involved in the problem to be optimized.

\subsection{Online Transformed MS-CSC Model}
The previous MS-CSC model is specifically designed for rain removal in a fixed length of video under the assumption that the rain is of consistent configuration along time. Specifically, the rain feature maps $\mathcal{M}$ (as defined in Eq. (\ref{rain})) of all video frames attained under fixed filters are assumed to follow a unique independent and identically distributed Laplacian. The real rain shapes, however, are always both correlated and distinctive along time, and varied from frame to frame across the entire video. The simple encoding manner of MS-CSC is thus inappropriate to real scenarios. We thus present the online MS-CSC model, which not only provides a more proper way to describe temporally dynamic rain/snow, but also makes the method more efficient and potentially applicable to streaming videos with continuously increasing frames in real time.

Denote the newly coming frame as $X^t\in R^{h\times w}$, where $h$ and $w$ represent the height and width of this frame, respectively, and $d=hw$ denotes the total number of pixels in this frame. Similar to (\ref{decomp}), we then decompose $X^t$ as the following three parts:
\begin{equation}\label{inti1}
X^t=B^t+F^t+R^t+E^t,
\end{equation}
where $B^t,R^t,F^t,E^t\in R^{h\times w}$ represent the background scene, rain layer, moving objects and background noise of the current frame, respectively. We then put forward the schemes to model these parts based on the dynamic characteristics of rain/snow.

\subsubsection{Modeling dynamic rain/snow layer}
Similar to the aforementioned offline MS-CSC model, we also adopt MS-CSC model~\cite{Zeiler10deconvolutional} to represent the the repetitive local patterns and multi-scale shapes of rain streaks, namely:
\begin{equation}
R^t=\sum\limits_{k=1}^K \sum\limits_{s=1}^{s_k} D_{ks}^t\otimes M_{ks}^t,
\end{equation}
where $M^t=\{M_{ks}^t\}_{k,s=1}^{K,s_k}\subset R^{h\times w}$ is a set of feature maps that approximate the rain streak positions, and  $D^t=\{D_{ks}^t\}_{k,s=1}^{K,s_k}\subset R^{p_k\times p_k}$ denotes the filters representing the repetitive local patterns of rain streaks. $K$ and $s_k$ denote the filter number and the filter at the $k$-th scale, respectively.

Similar to the MS-CSC model, we also assume the feature map $M_{ks}^t$ of the current frame $X^t$ follows a Laplacian distribution (i.e., imposed with $L_1$ penalty as Eq. (\ref{mcsc}), which, however, has its specific scale parameter $b_{ks}^t$ different with others, namely:
\begin{align}
M_{ks}^t \sim \mathrm{Laplacian}(M_{ks}^t|0,b_{ks}^t),
\end{align}
where the scale parameter $b_{ks}^t > 0$ is specified for the current frame reflecting the specific rain shape in this frame. Furthermore, the correlation of rain between current and previous frames is represented by the following prior term imposed on $b_{ks}^t$:
\begin{align}
b_{ks}^t\sim \ {\mathrm{Inv\text{-}Gam}}(b_{ks}^t|N^{t-1}-1, N^{t-1}b_{ks}^{t-1}),
\end{align}
where $N^{t-1}=(t-1)d$ and $b_{ks}^{t-1}$ represent the scale parameter learned from the previous frames. Here $\mathrm{Inv\text{-}Gam}(\cdot)$ denotes the Inverse-Gamma distribution, a conjugate prior to $b_{ks}^t$, whose mode is exactly the one of previously learned (i.e., $b_{ks}^{t-1}$). It is then naturally delivered that the correlation of rain shapes between current frame and the learned knowledge from previous ones.

In the way as aforementioned, the dynamic characteristic of rain/snow across a video can then be rationally represented. In specific, the scale parameter in each frame is specifically learned and different from one another, finely representing the distinctiveness (i.e. 'non-identical') of rain/snow among different frames. Furthermore, the scale parameter of feature map distribution for the current frame is regularized by that of previously learned ones, well encoding the correlation (i.e., 'non-independent') across especially adjacent frames. The model is thus expected to better adapt to the variations of the dynamic rain/snow.

\subsubsection{Modeling moving object and background noise layers}
Following the MS-CSC model, we also adopt MRF to detect the moving objects. Let $H\in R^{h\times w}$ be a binary matrix denoting the moving object support, which is defined as
\begin{equation}
H_{ij}=
\begin{cases}
1, & \text{location} ~ (i,j) ~ \text{is moving objects},\\
0, & \text{location} ~ (i,j) ~ \text{is background}. \\
\end{cases}
\end{equation}
Let $H^\bot$ be complementary of $H$ satisfying $H+H^\bot=1$. Eq.(\ref{inti1}) can then be equivalently expressed as:
\begin{equation}\label{bg1}
X^t={H^t}^\bot\circ B^t+H^t\circ F^t+R^t+E^t.
\end{equation}
Like the optimization problem (\ref{mcsc}), by assuming all elements of the background noise $E^t$ follow a Gaussian distribution with zero mean and variance ${(\sigma^t)}^2$, we can then get the probabilistic model for the component $x_{ij}^t$ of $X^t$ as follows:
\begin{align}\label{likelihood1}
	x_{ij}^t \sim \ N(x_{ij}^t|({H_{ij}^t}^\bot \circ B_{ij}^t+ H_{ij}^t \circ F_{ij}^t+R_{ij}^t),{(\sigma^t)}^2).
\end{align}

Similar to the dynamic shapes of rain in practical video, the background noise embedded in the video is also with dynamic forms, and also both distinctive and correlated among video frames. We can then also represent this dynamic knowledge. Specifically, for video noise in the current frame with variance ${\sigma^t}^2$, we model it in the similar modeling manner as aforementioned, i.e., imposing conjugate prior to ${(\sigma^t)}^2$ as:
\begin{align}\label{prior}
	{(\sigma^t)}^2\sim  \ {\mathrm{Inv\text{-}Gam}}({(\sigma^t)}^2|\frac{N^{t-1}}{2}-1, \frac{N^{t-1}{(\sigma^{t-1})}^2}{2}),
\end{align}
where $N^{t-1}=(t-1)d$ and ${(\sigma^{t-1})}^2$ denote the variance of Gaussian noise learned from the previous frames. The mode of this prior is also the knowledge previously learned (i.e., ${(\sigma^{t-1})}^2$). This encoding manner is thus also able to deliver the dynamic property of noises/snow along the video.

\subsubsection{Modeling dynamic video background}
To tackle dynamic shapes of background scenes in a video due to camera jitter, i.e., video transformations like translation, rotation and scaling, a flexible affine transformation operation is imposed on the background. In the decomposition form (\ref{inti1}) for the current frame $X^t$, the background component $B^{t}$ is expressed to be transformed from the previous one $B^{t}$ as $B^t=B^{t-1}\odot\tau$, where $\tau$ denotes the transformed operator implemented on the initial background $B^{t-1}$, and can be formulated as an affine or projective
transformation~\cite{Yong2107Robust}. Then, Eq.(\ref{bg1}) and (\ref{likelihood1}) are reformulated as:
\begin{align}
X^t={H^t}^\bot\circ (B^{t-1}\odot\tau)+H^t\circ F^t+R^t+E^t.
\end{align}
\vspace{-0.8cm}
\begin{align}\label{tbg}
	&x_{ij}^t\!\sim\!\ N(x_{ij}^t|((H_{ij}^t)^\bot\!\circ\! (B_{ij}^{t-1}\odot\tau)\!\!+\! H_{ij}^t\!\circ\!F_{ij}^t\!+\!R_{ij}^t),{(\sigma^t)}^2).
\end{align}

\subsubsection{Online Transformed MS-CSC Model}
For convenience, we denote all involved parameters as $\Theta=\{H,\tau,D,M,F,\sigma^2,b\}$ and the parameters in the current and last frames as $\Theta^{t}$ and $\Theta^{t-1}$, respectively. Based on the models provided in the last sections, given the previous parameters $\Theta^{t-1}$ and newly coming frame $X^t$, we can then obtain the posterior distribution of $\Theta$ as follows:
\begin{align}
p(H^t,\tau,&D^t,M^t,F^t,{(\sigma^t)}^2,b^t|X^t,\Theta^{t-1})\nonumber\\
&\propto
p(X^t|H^t,\tau,F^t,D^t,M^t,{(\sigma^t)}^2)p({(\sigma^t)}^2|\Theta^{t-1})\nonumber\\
&p(M^t|b^t)p(b^t|\Theta^{t-1})p(H^t)p(D^t)p(F^t)p(\tau).
\end{align}
Through maximizing this posterior, the updated parameters $\Theta^{t}$ for the current frame can then be attained. This MAP problem can then be equivalently expressed as the following minimization problem:
\begin{align}\label{map1}
\mathcal{L}(\Theta^t)= &-\!\ln p(X^t|H^t\!,B^{t-1}\!,\tau,F^t\!,D^t\!,M^t\!,{(\!\sigma^t)}^2)\!+\!Q_E({(\!\sigma^t)}^2) \nonumber\\
&-\!\sum_{k,s}\!\ln p(M_{k,s}^t|b_{k,s}^t)+Q_R(b^t)+Q_F(F^t,H^t), \nonumber\\
s.t.& \quad R^t=\sum\limits_{k,s}D_{ks}^t\otimes M_{ks}^t,  \quad \parallel D_{ks}^t\parallel_F^2 \leq 1,
\end{align}
where
\begin{align}
& Q_E({(\sigma^t)}^2)=N^{t-1}(\ln\sigma^t+{(\sigma^{t-1})}^2/2{(\sigma^t)}^2),\\
& Q_R(b^t)=N^{t-1}\sum_{k,s}(\ln b_{ks}^t+b_{ks}^{t-1}/b_{ks}^t), \\
& Q_F(F^t,H^t)=\lambda\!\parallel\! F^t\! \parallel_{TV}\!+\alpha\!\parallel\! H^t\!\parallel_{\mathrm{3DTV}}\!+\!\beta\!\parallel\! H^t\!\parallel_1.
\end{align}
Specifically, $Q_E({(\sigma^t)}^2)$ and $Q_R(b^t)$ correspond to the regularization terms for the distributions of feature map $M_{ks}^t$ and noises embedded in $x$, respectively, which can be more intuitively understood by the following equivalent forms:
\begin{align}\label{kl1}
Q_E({(\sigma^t)}^2)=N^{t\!-\!1}\!D_{\mathrm{KL}}(N(x|0,{(\sigma^{t-1})}^2)\!\parallel \! N(x|0,{(\sigma^t)}^2)),
\end{align}
\vspace{-0.5cm}
\begin{align}\label{kl2}
Q_R(b^t)=N^{t\!-\!1}\!\sum_{k,s}\!D_{\mathrm{KL}}(L(M_{ks}^t\!|0,b_{ks}^{t-1}\!)\!\parallel\! L(M_{ks}^t\!|0,b_{ks}^t\!))
\vspace{-0.5cm}
\end{align}
where $D_{\mathrm{KL}}(\cdot\parallel\cdot)$ denotes the KL divergence between two distributions. Particularly, it can be easily observed that $Q_R(b^t)$ functions to rectify the rain streaks on the current frame with parameter $b_{ks}^t$ to approximate the previously learned rain streaks with parameter $b_{ks}^{t-1}$, so as to make the rain shapes in the adjacent frames correlated. Similarly, the regularization term $Q_E({(\sigma^t)}^2)$ inclines to enforce the background noise in the current frame close to that embedded in the previous ones. This easily explains why our method can fit dynamic rain, as well as varying background noises, in a video with evidently non-i.i.d. configurations.

The corresponding augmented Lagrangian function of Eq. (\ref{map1}) can be written as follows:
\begin{align}\label{Lagrange1}
\mathcal{L}&(\Theta^t)\!=\!\frac{1}{2(\sigma^t)^2}\!\parallel\! X^t\!-\!{(\!H^t)}^\bot\!\circ\!(\!B^{t-1}\!\odot\!\tau\!)\!-\!H^t\!\circ\! F^t\!-\!R^t\!\parallel_F^2\! \nonumber\\
&+d\!\ln\!\sigma\!+\!N^{t-1}(\ln\sigma^t\!+\!\frac{{\sigma^{t-1}}^2}{2{\sigma^t}^2})\!+\!\alpha\!\parallel \!H^t\!\parallel_{\mathrm{3DTV}}+\beta\parallel\! H^t\!\parallel_1  \nonumber\\
&+\sum\limits_{k,s}(d\ln b_{ks}^t\!+\!\frac{1}{b_{ks}^t}\parallel \!M_{ks}^t\!\parallel_1)\!+\!\sum\limits_{k,s}N^{t-1}(\ln b_{ks}^t\!+\!\frac{b_{ks}^{t-1}}{b_{ks}^t}\!) \nonumber\\
&+ \lambda \parallel\! F^t\!\parallel_{TV}+\frac{\rho}{2}\parallel \sum\limits_{k,s}D_{ks}^t\otimes M_{ks}^t\!-\!R^t\!+\!T^t\parallel_F^2,
\end{align}
where $T^t$ and $\rho$ are the Lagrange variable and the penalty parameter, respectively.

\subsection{ADMM Algorithm}
We can then readily adopt ADMM algorithm to iteratively optimize each variable involved in Eq. (\ref{Lagrange1}). To simplify the relevant subproblems, we will utilize the following equation:
\begin{align}
	&\parallel\! X^t\!-\!((H^t)^\bot\!\circ(B^{t-1}\odot\tau)\!+\!H^t\!\circ\! F^t\!+\!R^t)\parallel_F^2 \ = \ \nonumber\\
	&\parallel\! (\!H^t)^\bot\!\circ\!(\!X^t\!-\!(\!B^{t-1}\odot\tau)\!-\!R^t)\!\parallel_F^2\!+\!\parallel\! H^t\!\circ\!(\!X^t\!-\!F^t\!-\!R^t)\!\parallel_F^2.\nonumber
\end{align}
Next, we discuss how to solve each subproblem separately.

\textbf{\textit{Update}} $H^t$: The subproblem with respect $H^t$ is
\begin{align}
	\min_{H^t}\ &\frac{1}{2(\sigma^t)^2}\parallel X^t-(H^t)^\bot\circ(B^{t-1}\odot\tau)-H^t\circ F^t-\!R^t\!\parallel_F^2 \nonumber\\
	&+\alpha \parallel H^t\parallel_{3DTV}+\beta\parallel H^t\parallel_1.
	\label{H}
\end{align}
This subproblem is a standard energy minimization problem, which can be efficiently solved by graph cut algorithm \cite{boykov2001fast,kolmogorov2004energy}.

\textbf{\textit{Update}} $F^t$: The subproblem with respect to $F^t$ is
\begin{equation}
	\min_{F^t}\ \parallel H^t\circ(X^t-F^t-R^t)\parallel_F^2 + 2(\sigma^t)^2\lambda\parallel F^t \parallel_{TV},
	\label{F}
\end{equation}
which is easily solved by the TV regularization algorithm \cite{wang2014highly}.

\textbf{\textit{Update}} $\tau$ and $B^t$: Since $B^{t-1}\odot\tau$ is a nonlinear geometric transform, it's hard to directly optimize it and we resort to the following linear approximation:
\begin{align}\label{onlineB}
B^t=B^{t-1}\odot\tau+J\triangle\tau,
\end{align}
where $J$ is the Jacobian of $X^t$ with respect to $\tau$. We can iteratively approximate the original nonlinear transformation with a locally linear approximation, as $\tau = \tau+\triangle\tau$. Therefore, the subproblem with respect to $\tau$ can be reformulated as:
\begin{align}
\min_{\triangle\tau}\parallel\!({H^t})^\bot\circ (X^t\!-\! B^{t-1}\odot\tau-J\triangle\tau-\! R^t)\!\parallel^2.
\end{align}
It can be solved in closed-form. The solution is:
\begin{align}\label{vtau}
\bigtriangleup\tau = (J'J)^{-1}J'(X^t-R^t-B^{t-1}\odot\tau).
\end{align}
Fixing $\bigtriangleup\tau$, we can use Eq. (\ref{onlineB}) to update the background.

\textbf{\textit{Update}} $M^t$: The subproblem with respect $M^t$ is
\begin{equation}\label{M}
	\!\min_{M_{ks}^t} \!\frac{1}{2}\!\parallel\! \sum\limits_{k=1}^K\! \sum\limits_{s=1}^{s_k}\! D_{ks}^t\!\otimes\! M_{ks}^t\!-\!R^t\!+\!T^t\! \parallel\!_F^2\!+\!\sum\limits_{k=1}^K\! \sum\limits_{s=1}^{s_k}\! \frac{b_{ks}^t}{\rho}\!\parallel \!M_{ks}^t\!\parallel\!_1.
\end{equation}
This subproblem is a standard CSC problem and can be readily solved by \cite{Wohlberg14CSC}, which adopts the ADMM scheme and FFT to improve computation efficiency.

\textbf{\textit{Update}} $D^t$: The subproblem with respect to $D^t$ is
\begin{equation}\label{D}
	\min\limits_{D^t} \!\frac{1}{2}\!\parallel\! \sum\limits_{k=1}^K\! \sum\limits_{s=1}^{s_k}\! D_{ks}^t\!\otimes\! M_{ks}^t\! -\!R^t\!+\!T^t\! \parallel_F^2\!,
	~~\text{s.t.} \!\parallel\! D_{ks}^t\!\parallel_F^2\! \leq\! 1.
\end{equation}
We use online learning algorithm for sparse coding~\cite{Mairal2009Online} to update the filters $D^t\!=\!\{D_{ks}^t\}_{k,s=1}^{K,n_k}$. The algorithm utilizes block-coordinate descent with warm restarts
$D^{t-1}=\{D_{ks}^{t-1}\}_{k,s=1}^{K,n_k}$.

\textbf{\textit{Update}} $R^t$: The subproblem with respect to $R^t$ is
\begin{align}
	\min_{R^t}\ &\!\frac{1}{2(\sigma^t)^2}\parallel X^t-(H^t)^\bot\circ(B^{t-1}\odot\tau)\!-H^t\circ F^t-\!R^t\parallel_F^2 \nonumber\\
	&+\frac{\rho}{2} \parallel \sum\limits_{k=1}^K \sum\limits_{s=1}^{s_k} D_{ks}^t\otimes M_{ks}^t - R^t + T^t \parallel_F^2.
\end{align}
The closed-form solution is
\vspace{-1mm}
\begin{equation}\label{R}
\vspace{-1mm}
	R^t=(X^t\!-\Gamma^t)/(1+\rho(\sigma^t)^2)
\end{equation}
where $\Gamma^t=(H^t)^\bot\circ(B^{t-1}\odot\tau)\!+\!H^t\circ F^t\!-\!\rho(\sigma^t)^2(\sum\limits_{k,s}^{K, s_k} \!D_{ks}^t\!\otimes \!M_{ks}^t\!+\!T^t)$.

\textbf{\textit{Update}} $T^t$: Following the general ADMM setting, $T^t$ can be updated as:
\vspace{-0.7mm}
\begin{equation}\label{T}
\vspace{-0.7mm}
	T^t =T^{t-1} + \sum_{k,s} D_{ks}^t\otimes M_{ks}^t-R^t.
\end{equation}

\textbf{\textit{Update}} $(\sigma^t)^2$: The subproblem with respect $(\sigma^t)^2$ is
\vspace{-0.7mm}
\begin{align}\label{sigma}
\vspace{-0.7mm}
	\min_{(\sigma^t)^2}\ &\frac{1}{2(\sigma^t)^2}\parallel X^t-( (H^t)^\bot\circ B^t+ H^t\circ F^t +R^t)\parallel_F^2 \nonumber\\
	&+d\ln\sigma^t+N^{t-1}(\ln\sigma^t+\frac{{\sigma^{t-1}}^2}{2(\sigma^t)^2}).
\end{align}
Its closed-form solution is:
\vspace{-0.7mm}
\begin{align}\label{sigma1}
\vspace{-0.7mm}
	(\sigma^t)^2=\frac{1}{t}(\overline{\sigma}^t)^2+\frac{t-1}{t}{\sigma^{t-1}}^2,
\end{align}
where $(\overline{\sigma}^t)^2=\frac{1}{d}\parallel X^t-( (H^t)^\bot\circ B^t+ H^t\circ F^t +R^t)\parallel_F^2$.

\vspace{+2mm}
\textbf{\textit{Update}} $b_{ks}^t$: The subproblem with respect to $b_{ks}^t$ is
\begin{align}\label{b}
	\!\min_{b_{ks}^t}\ \!(d\!+\!N^{t-1})\!\ln\! b_{ks}^t\!+\!(b^{t}_{ks})^{-1}(\parallel\! M_{ks}^t\!\parallel_1\!+N^{t-1}b_{ks}^{t-1}).
\end{align}
Its closed-form solution is:
\begin{align}\label{b1}
	b_{ks}^t=\frac{1}{t}\overline{b}_{ks}^t+\frac{t-1}{t}b_{ks}^{t-1},
\end{align}
where $\overline{b}_{ks}^t=\frac{1}{d}\parallel\! M_{ks}^t\!\parallel_1$.

The algorithm for solving this online transformed MS-CSC (OTMS-CSC) model can then be summarized as Algorithm~\ref{alg2}.

\begin{algorithm}
	\caption{Algorithm for OTMS-CSC Model}\label{alg2}
	\begin{algorithmic}[1]
		\renewcommand{\algorithmicrequire}{\textbf{Input:}}
		\renewcommand{\algorithmicensure}{\textbf{End}}
		\REQUIRE The newly coming frame: {\small $X^t\in\mathbb{R}^{h\times w}$}; model variables of last frame: {\small $\Theta^{t-1}=\{H^{t-1}, B^{t-1}, D^{t-1}\}$}; the parameters of last frame: {\small $\{{(\sigma^{t-1})}^2, b^{t-1}\}$}.
		\renewcommand{\algorithmicrequire}{\textbf{Initialization:}}
		\renewcommand{\algorithmicensure}{\textbf{End}}
		\REQUIRE {\small $\{H^t, D^t\}=\{H^{t-1}, D^{t-1}\}$}, $\tau=0$.
		
		\IF {$t/l == 0$ }
		\STATE
		update {\small $B^{t-1}=\hat{B}^{t-1}$} by using the strategy suggested in Sec. 3.4.2.
		\ENDIF
		\WHILE {not converge}
		\STATE
		Update {\small $\triangle\tau$} by Eq. (\ref{vtau}) and update {\small $\tau=\tau+\triangle\tau$}.
		\STATE
		Update aligned background {\small $B^t$} by Eq. (\ref{onlineB}).
		\STATE
		Update {\small $H^t,F^t$}  by Eq.(\ref{H}), (\ref{F}), respectively.
		\STATE
		Update {\small $M^t, D^t$} by Eq.(\ref{M}), (\ref{D}), respectively.
		\STATE
		Update {\small $R^t, T^t$} by Eq.(\ref{R}), (\ref{T}), respectively.
		\STATE
		Update $(\sigma^t)^2, b^t$ by Eq.(\ref{sigma1}), (\ref{b1}), respectively.
		
		\ENDWHILE
		\renewcommand{\algorithmicrequire}{\textbf{Output:}}
		\renewcommand{\algorithmicensure}{\textbf{End}}
		\REQUIRE {\small $\Theta^t=\{H^{t}, D^{t}, B^{t}, F^{t},{\sigma^{t}}^2, b^{t}\}$};\\
		\qquad  Recovered frame = {\small ${H^t}^{\perp}\circ B^t+ H^t\circ F^t$}.
	\end{algorithmic}
\end{algorithm}
\vspace{-2mm}

\subsection{Some Remarks}
\subsubsection{Explanation for function of $D_{KL}$ regularizations}
It should be noted that the $D_{KL}$ regularization in Eq. (\ref{kl1}) and Eq. (\ref{kl2}) intrinsically conduct the superiority of the proposed OTMS-CSC model for removing dynamic rain/snow. Specifically, the offline MS-CSC model \cite{Li18Multiscale} intrinsically specifies one unique value for the parameter $\sigma^2$ as well as $b$ to represent the background noise variance and scale parameter in feature map representing rain/snow, respectively, for all the frames of the video. The offline model is thus only suitable to be used in the video with static background and consistent rain/snow shapes. The OTMS-CSC model, however, can finely handle dynamic rain with videos with dynamic rain and varying background noises. This advantage is naturally conducted by the fact that the model assumes that each frame has its own specific noise parameter $(\sigma^{t})^2$ and scale parameter $b^{t}$, by simultaneously fitting the knowledge of the current frame and being regularized by those ($(\sigma^{t-1})^{2}$ and $b^{t-1}$) obtained from the previous frames. This makes this model, implemented for each new frame in an online mode, better adapt the specific structures of rain/snow or background for the current frame, generally varied from those for previous ones.

\begin{figure}[tb]\vspace{-4mm}
	\begin{minipage}{1\linewidth}
		\centerline{\includegraphics[width=1\linewidth]{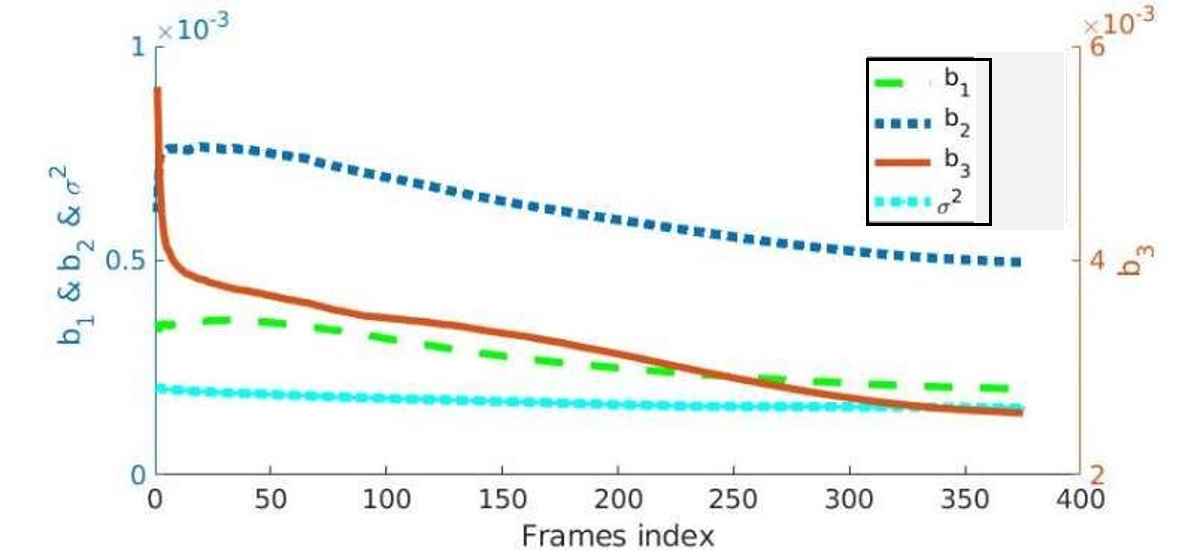}}
	\end{minipage}\vspace{-2mm}
	\caption{The changing tendency of the noise variance $(\sigma^t)^2$ and the scale parameter $b^t$ along a video (as shown in Fig. \ref{fig:online}) containing dynamic snow varying from heavy to light.
Since there are three different scales of filters (used for $13\times 13$, $9\times 9$, $3\times 3$ patch sizes , respectively) are utilized, there are three scale parameter changing curves.
}
	\label{fig:kl}
	\vspace{-5mm}
\end{figure}

To more intuitively clarify this point, we illustrate in Fig.~\ref{fig:kl} the changing tendencies of parameters $(\sigma^t)^2$ and $b^t$ for a sequence of video frames, containing snow varying from heavy to light, as shown in Fig.~\ref{fig:light}. It can be seen that both $(\sigma^t)^2$ and $b^t$ are gradually decreasing along time, finely reflecting the dynamic changes of snow along time.

\subsubsection{Background Amelioration}
Our method gradually updates the background $B^t$ of the current frame from the affine transformation on that of the last frame $B^{t-1}$ by Eq. (\ref{onlineB}). Due to constantly temporal scene shifting of the videos (especially brought by the camera moving along a certain direction in a short time) and incremental accumulation of computing errors, the recovered video background tends to be gradually deviated from the real one, which always makes the rain-removed videos look more or less blurry after a period of algorithm computing. 
To alleviate this issue, our algorithm needs to specifically ameliorate the background knowledge $B^t$ after implementing certain frames by our algorithm.

Our strategy is as follows: When our algorithm is run $l$ iterations (the current frame is denoted as the $t^{th}$ one), we then pick up two frames before and after current frame to get a subgroup as:
\begin{align}
	\mathcal{\hat{X}}^t=[X^{t-2},X^{t-1},X^t,X^{t+1},X^{t+2}].
\end{align}
We then easily align all other frames under the reference of the current frame by using the similar manner as we introduced in Eq. (\ref{onlineB}), to obtain the aligned subgroup as (a $h\times w\times 5$ tensor):
\begin{align}
\mathcal{T\hat{X}}=[TX^{t-2},TX^{t-1},X^t,TX^{t+1},TX^{t+2}],
\end{align}
where $TX^{j}=X^t\odot\tau^j$ ($j=t-2,t-1,t+1,t+2$), and $\tau^j$ is calculated readily by Eq. (\ref{onlineB})-(\ref{vtau}). Then we can easily calculate the optimal rank-one approximation $\hat{B}^{t-1}$ of the unfolded matrix $T\hat{X}\in R^{hw\times 5}$ of $\mathcal{T\hat{X}}$ efficiently by SVD, and replace $B^{t-1}$ as $\hat{B}^{t-1}$ to get the new ameliorated background initialization.
\begin{figure*}[!htb]\vspace{-2mm}
\begin{minipage}{0.24\linewidth}
  \centerline{\includegraphics[width=1\linewidth]{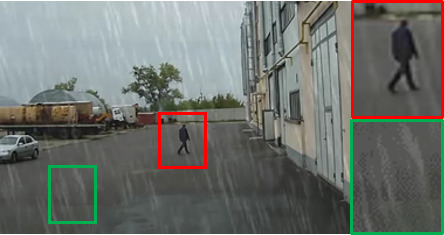}}
\end{minipage}
\hfill
\begin{minipage}{.24\linewidth}
  \centerline{\includegraphics[width=1\linewidth]{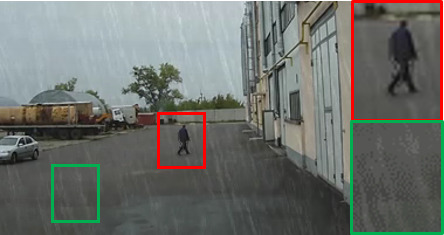}}
\end{minipage}
\hfill
\begin{minipage}{.24\linewidth}
  \centerline{\includegraphics[width=1\linewidth]{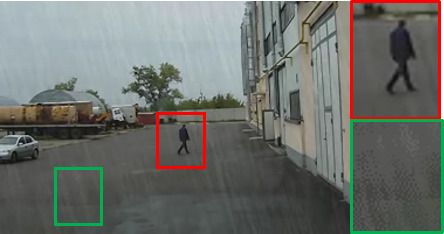}}
\end{minipage}
\hfill
\begin{minipage}{0.24\linewidth}
  \centerline{\includegraphics[width=1\linewidth]{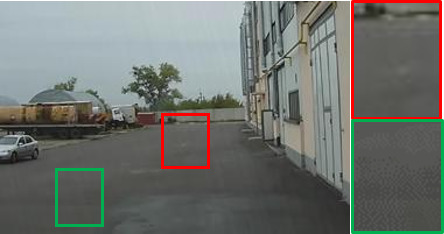}}
\end{minipage}
\vfill
\begin{minipage}{0.24\linewidth}
  \centerline{\includegraphics[width=1\linewidth]{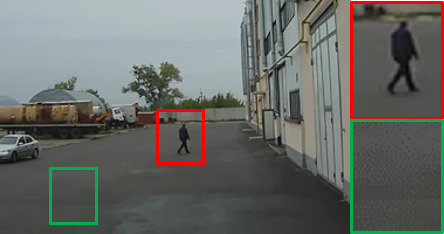}}
  \vspace{-1mm}
  \centerline{\small{(a) Input/GT }}
\end{minipage}
\hfill
\begin{minipage}{.24\linewidth}
  \centerline{\includegraphics[width=1\linewidth]{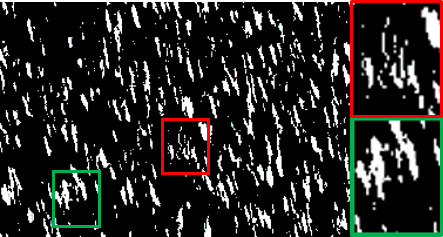}}
  \vspace{-1mm}
  \centerline{\small{(b)  Garg et al. \cite{Garg2007Vision}}}
\end{minipage}
\hfill
\begin{minipage}{.24\linewidth}
  \centerline{\includegraphics[width=1\linewidth]{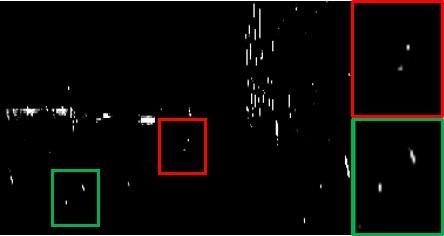}}
  \vspace{-1mm}
  \centerline{\small{(c)  Jiang et al. \cite{jiang2017cvpr}}}
\end{minipage}
\hfill
\begin{minipage}{0.24\linewidth}
  \centerline{\includegraphics[width=1\linewidth]{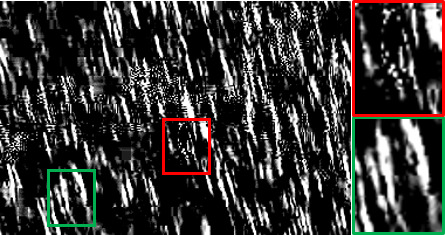}}
  \vspace{-1mm}
  \centerline{\small{(d) Ren et al. \cite{ren2017cvpr}}}
\end{minipage}
\vfill
\begin{minipage}{0.24\linewidth}
  \centerline{\includegraphics[width=1\linewidth]{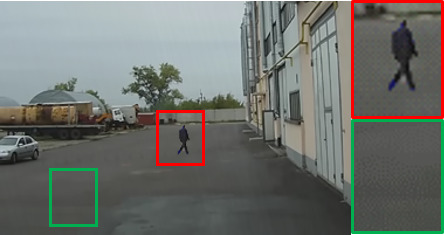}}
\end{minipage}
\hfill
\begin{minipage}{0.24\linewidth}
  \centerline{\includegraphics[width=1\linewidth]{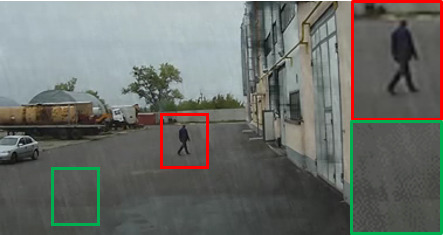}}
\end{minipage}
\hfill
\begin{minipage}{0.24\linewidth}
  \centerline{\includegraphics[width=1\linewidth]{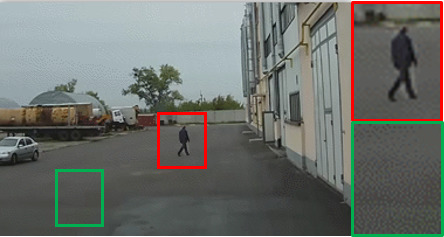}}
\end{minipage}
\hfill
\begin{minipage}{0.24\linewidth}
  \centerline{\includegraphics[width=1\linewidth]{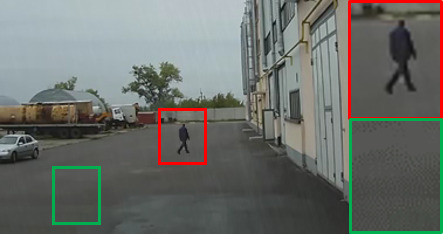}}
\end{minipage}
\vfill
\begin{minipage}{0.24\linewidth}
  \centerline{\includegraphics[width=1\linewidth]{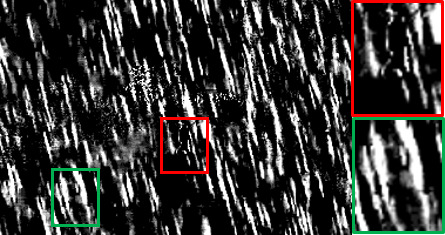}}
  \vspace{-1mm}
  \centerline{\small{(e) Wei et al. \cite{wei2017should}}}
\end{minipage}
\hfill
\begin{minipage}{0.24\linewidth}
  \centerline{\includegraphics[width=1\linewidth]{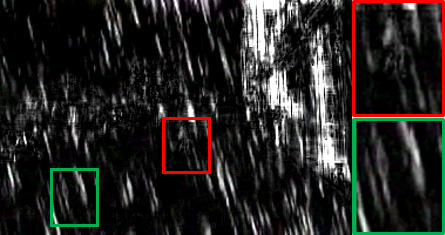}}
  \vspace{-1mm}
  \centerline{\small{(f) Liu et al. \cite{Liu18Erase} }}
\end{minipage}
\hfill
\begin{minipage}{0.24\linewidth}
  \centerline{\includegraphics[width=1\linewidth]{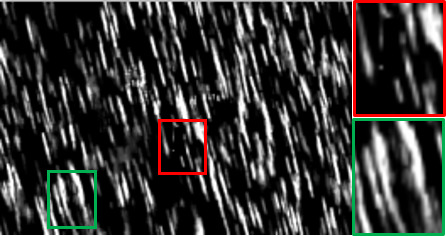}}
  \vspace{-1mm}
  \centerline{\small{(g) MS-CSC}}
\end{minipage}
\hfill
\begin{minipage}{0.24\linewidth}
  \centerline{\includegraphics[width=1\linewidth]{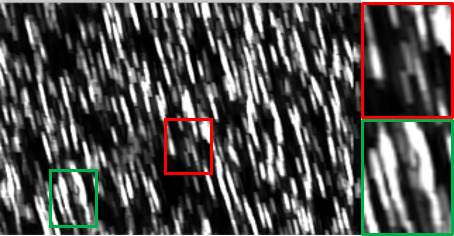}}
  \vspace{-1mm}
  \centerline{\small{(h) OTMS-CSC}}
\end{minipage}
\caption{(a) An input rainy frame (upper) and its groudtruth clean one (lower). (b)-(h) Recovered frames (upper) and extracted rain layers (lower) by different competing methods.}
\label{fig:park}\vspace{-2mm}
\end{figure*}

\subsubsection{Potential to be used for streaming videos}
It is evident that the proposed OTMS-CSC algorithm is implemented in an online mode, i.e., each time run on a unique newly coming frame. This learning manner makes our method potentially applicable to practical streaming videos. In specific, in each implementation stage for a frame $X^t$, the algorithm only requires a fixed memory to restore related parameters $H^{t}, B^{t}, D^{t}, {(\sigma^{t})}^2, b^{t}$. Besides, since the implementation is similar to each new frame, its time complexity is also fixed in each learning stage. This makes our method potentially feasible to the practical videos continuously coming with streaming format beyond current offline methods, which not only need increasingly more space complexity for larger length of videos, but also require increasingly larger time complexity for larger video sequence (even need to pre-implement the algorithms on the entire video again). This makes them hardly useable to this typical real video format in practice. Comparatively, our method makes the real-time execution of rain removal possible to be realized for practical streaming video. What we need to do is to improve the efficiency of our algorithm on one frame to make it gradually meet the real-time requirements. Possible regimes include further improvement on hardware power, further speed-up on algorithm implementation (like modify it distributed/parrallel or transform it in faster implementation platform), or replace some of its stages with faster algorithms. This is a meaningful issue worthy of making further endeavors in future research.

\begin{figure*}[!htb]
\begin{minipage}{0.32\linewidth}
  \centerline{\includegraphics[width=1\linewidth]{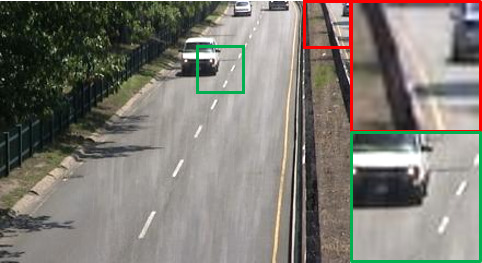}}
  \vspace{-1mm}
  \centerline{\small{(a) Input}}
\end{minipage}
\hfill
\begin{minipage}{0.32\linewidth}
	\centerline{\includegraphics[width=1\linewidth]{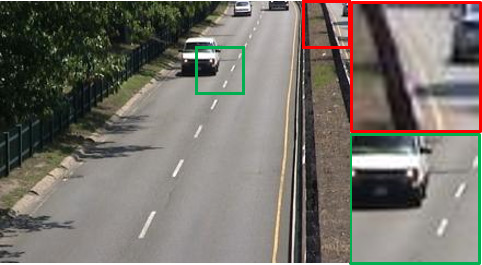}}
	\vspace{-1mm}
	\centerline{\small{(b) GT }}
\end{minipage}
\hfill
\begin{minipage}{.32\linewidth}
  \centerline{\includegraphics[width=1\linewidth]{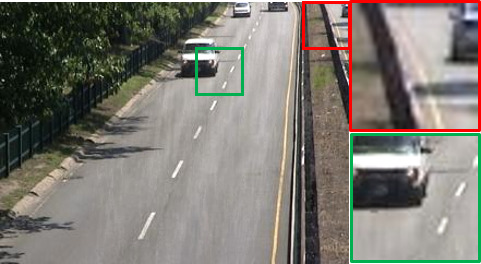}}
  \vspace{-1mm}
  \centerline{\small{(c)  Garg et al. \cite{Garg2007Vision}}}
\end{minipage}
\hfill
\begin{minipage}{.32\linewidth}
  \centerline{\includegraphics[width=1\linewidth]{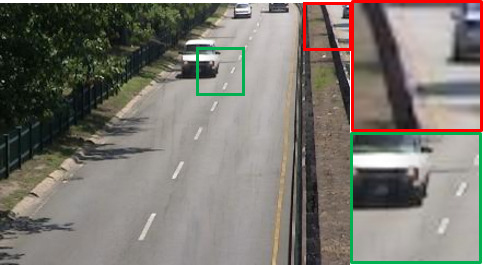}}
  \vspace{-1mm}
  \centerline{\small{(d)  Jiang et al. \cite{jiang2017cvpr}}}
\end{minipage}
\hfill
\begin{minipage}{0.32\linewidth}
  \centerline{\includegraphics[width=1\linewidth]{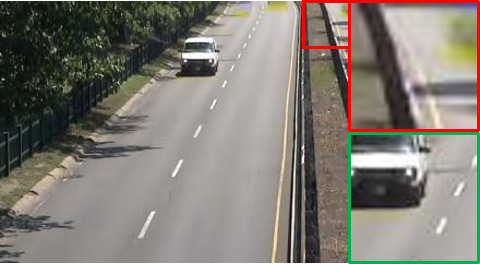}}
  \vspace{-1mm}
  \centerline{\small{(e) Ren et al. \cite{ren2017cvpr}}}
\end{minipage}
\hfill
\begin{minipage}{0.32\linewidth}
  \centerline{\includegraphics[width=1\linewidth]{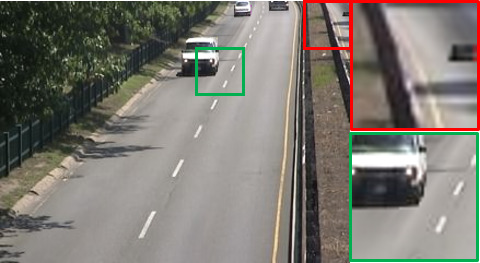}}
  \vspace{-1mm}
  \centerline{\small{(f) Wei et al. \cite{wei2017should}}}
\end{minipage}
\hfill
\begin{minipage}{0.32\linewidth}
  \centerline{\includegraphics[width=1\linewidth]{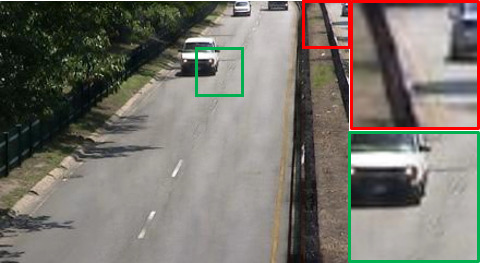}}
  \vspace{-1mm}
  \centerline{\small{(g) Liu et al. \cite{Liu18Erase}}}
\end{minipage}
\hfill
\begin{minipage}{0.32\linewidth}
  \centerline{\includegraphics[width=1\linewidth]{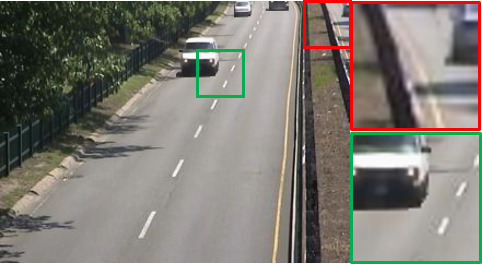}}
  \vspace{-1mm}
  \centerline{\small{(h) MS-CSC}}
\end{minipage}
\hfill
\begin{minipage}{0.32\linewidth}
  \centerline{\includegraphics[width=1\linewidth]{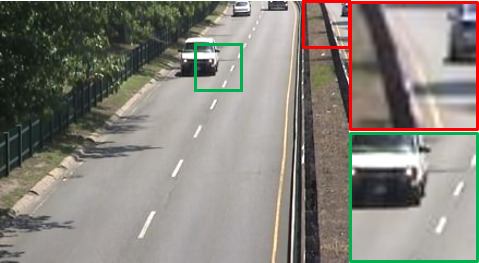}}
  \vspace{-1mm}
  \centerline{\small{(i) OTMS-CSC}}
\end{minipage}
\caption{(a)(b) An input frame with heavy rain and its groundtruth clean one. (c)-(i) Recovered frames obtained by different competing methods.
}
\label{fig:highway}\vspace{-5mm}
\end{figure*}

\begin{figure*}[!htb]
\vspace{-2mm}\begin{minipage}{0.242\linewidth}
  \centerline{\includegraphics[width=1\linewidth]{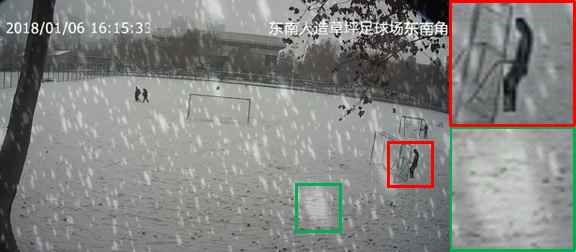}}
\end{minipage}
\hfill
\begin{minipage}{.242\linewidth}
  \centerline{\includegraphics[width=1\linewidth]{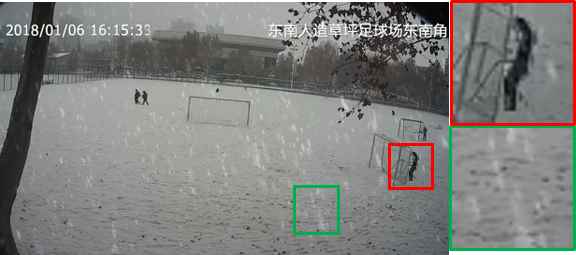}}
\end{minipage}
\hfill
\begin{minipage}{.242\linewidth}
  \centerline{\includegraphics[width=1\linewidth]{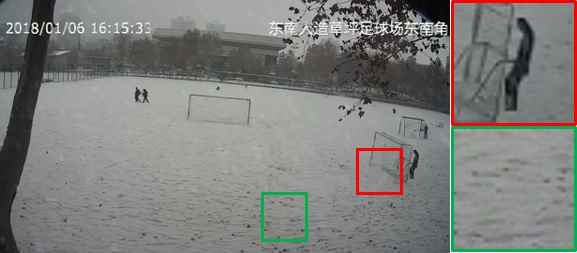}}
\end{minipage}
\hfill
\begin{minipage}{0.242\linewidth}
  \centerline{\includegraphics[width=1\linewidth]{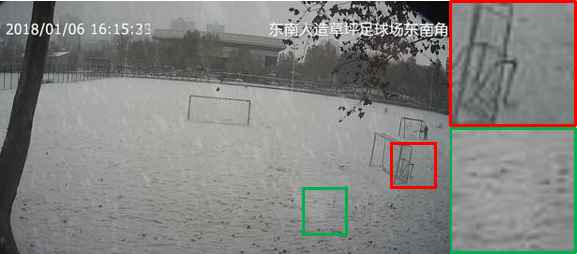}}
\end{minipage}
\vfill
\begin{minipage}{0.242\linewidth}
  \centerline{\includegraphics[width=1\linewidth]{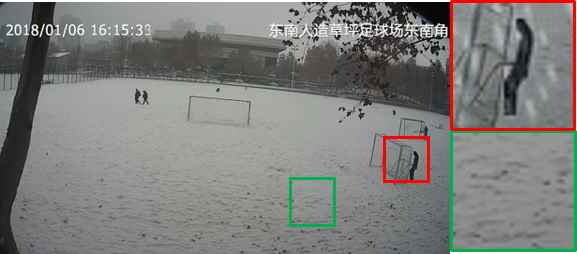}}
  \vspace{-1mm}
  \centerline{\small{(a) Input/GT }}
\end{minipage}
\hfill
\begin{minipage}{.242\linewidth}
  \centerline{\includegraphics[width=1\linewidth]{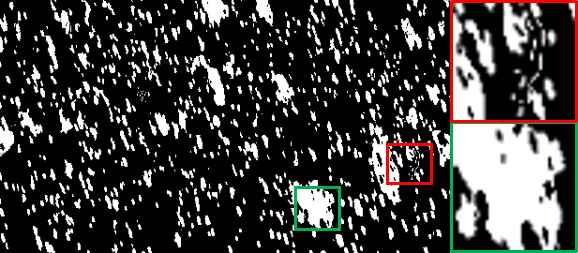}}
  \vspace{-1mm}
  \centerline{\small{(b)  Garg et al. \cite{Garg2007Vision}}}
\end{minipage}
\hfill
\begin{minipage}{.242\linewidth}
  \centerline{\includegraphics[width=1\linewidth]{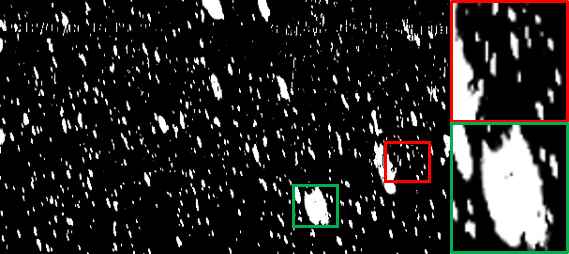}}
  \vspace{-1mm}
  \centerline{\small{(c)  Jiang et al. \cite{jiang2017cvpr}}}
\end{minipage}
\hfill
\begin{minipage}{0.242\linewidth}
  \centerline{\includegraphics[width=1\linewidth]{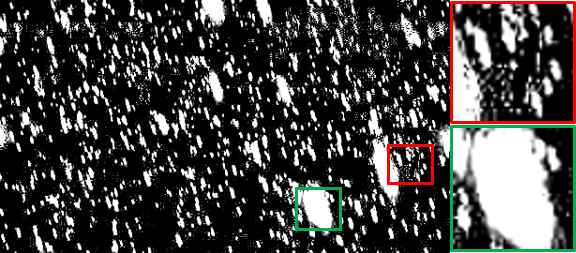}}
  \vspace{-1mm}
  \centerline{\small{(d) Ren et al. \cite{ren2017cvpr}}}
\end{minipage}
\hfill
\begin{minipage}{0.242\linewidth}
  \centerline{\includegraphics[width=1\linewidth]{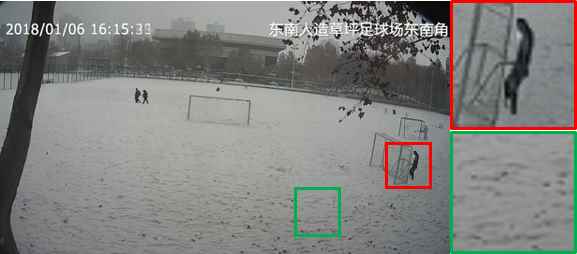}}
\end{minipage}
\hfill
\begin{minipage}{0.242\linewidth}
  \centerline{\includegraphics[width=1\linewidth]{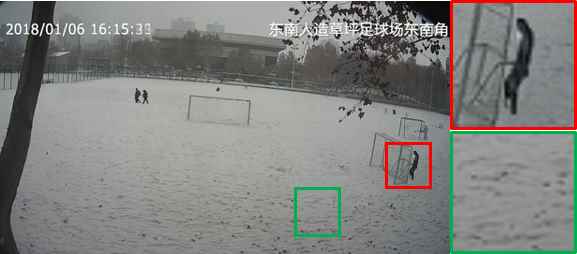}}
\end{minipage}
\hfill
\begin{minipage}{0.242\linewidth}
  \centerline{\includegraphics[width=1\linewidth]{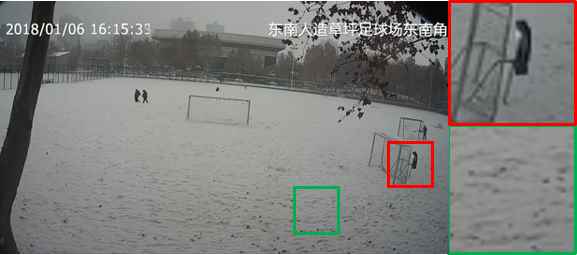}}
\end{minipage}
\hfill
\begin{minipage}{0.242\linewidth}
  \centerline{\includegraphics[width=1\linewidth]{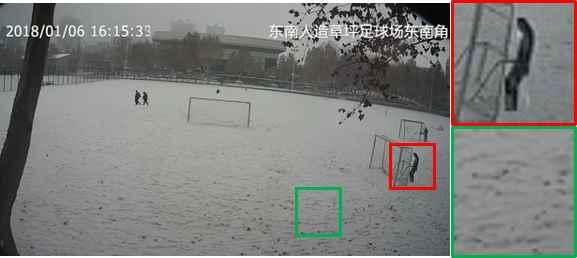}}
\end{minipage}
\hfill
\begin{minipage}{0.242\linewidth}
  \centerline{\includegraphics[width=1\linewidth]{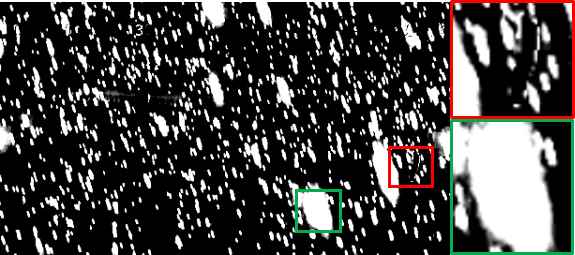}}
  \vspace{-1mm}
  \centerline{\small{(e) Wei et al. \cite{wei2017should}}}
\end{minipage}
\hfill
\begin{minipage}{0.242\linewidth}
  \centerline{\includegraphics[width=1\linewidth]{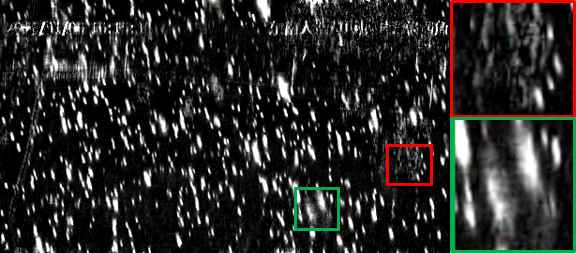}}
  \vspace{-1mm}
  \centerline{\small{(f) Liu et al. \cite{Liu18Erase}}}
\end{minipage}
\hfill
\begin{minipage}{0.242\linewidth}
  \centerline{\includegraphics[width=1\linewidth]{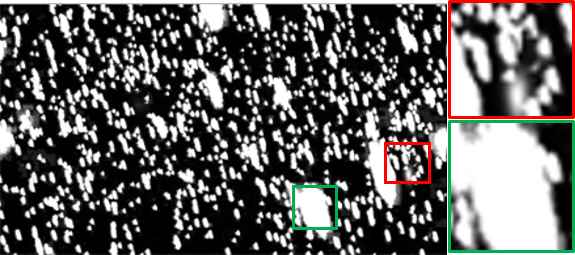}}
  \vspace{-1mm}
  \centerline{\small{(f) MS-CSC}}
\end{minipage}
\hfill
\begin{minipage}{0.242\linewidth}
  \centerline{\includegraphics[width=1\linewidth]{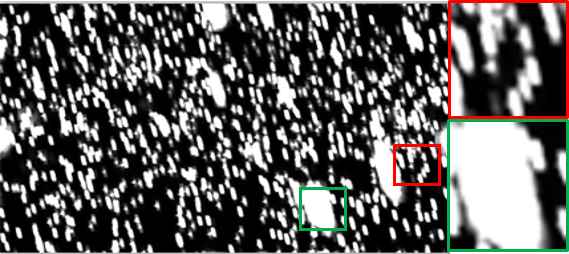}}
  \vspace{-1mm}
  \centerline{\small{(g) OTMS-CSC}}
\end{minipage}
\caption{
(a) an input snowy frame (upper) and its groudtruth clean one (lower). (b)-(g) Recovered frames (upper) and extracted snow layers (lower) by different competing methods.
}
\label{fig:4080}
\vspace{-2mm}
\end{figure*}

\begin{figure*}[!htb]
	\begin{minipage}{0.242\linewidth}
		\centerline{\includegraphics[width=1\linewidth]{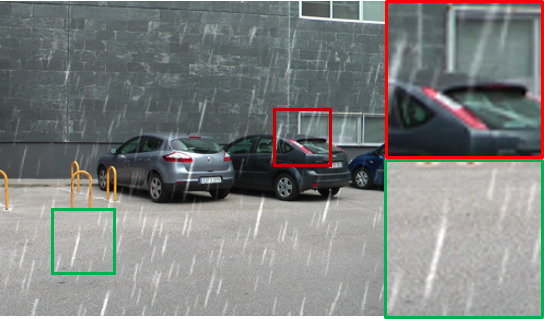}}
	\end{minipage}
	\hfill
	\begin{minipage}{0.242\linewidth}
		\centerline{\includegraphics[width=1\linewidth]{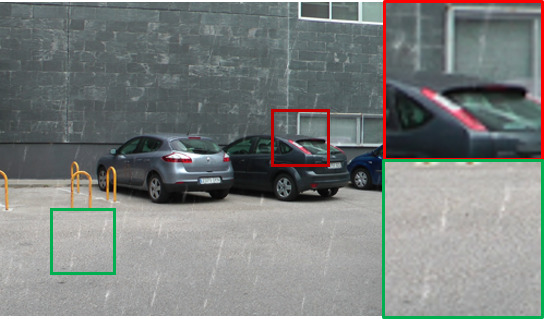}}
	\end{minipage}
	\hfill
	\begin{minipage}{.242\linewidth}
		\centerline{\includegraphics[width=1\linewidth]{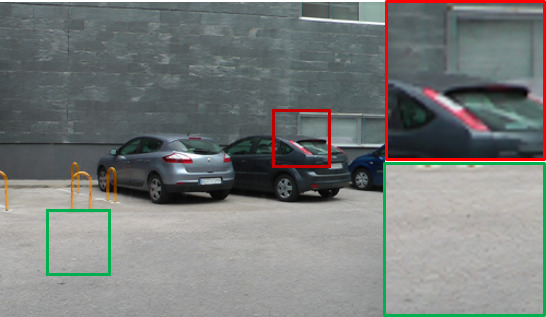}}
	\end{minipage}
	\hfill
	\begin{minipage}{0.242\linewidth}
		\centerline{\includegraphics[width=1\linewidth]{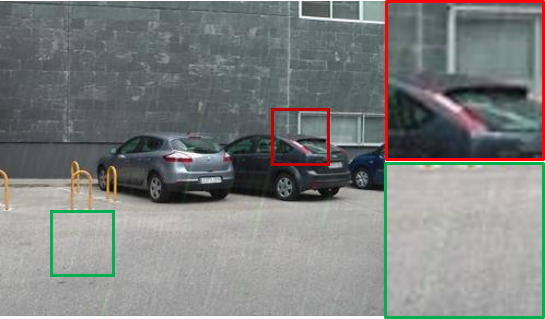}}
	\end{minipage}
    \vfill
	\begin{minipage}{.242\linewidth}
		\centerline{\includegraphics[width=1\linewidth]{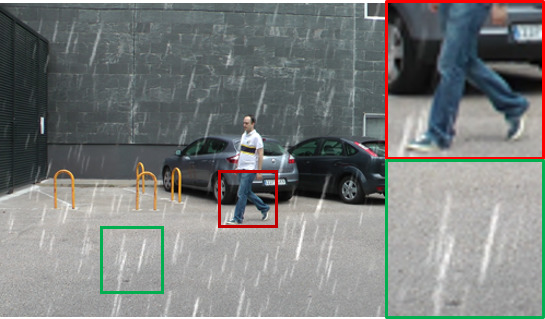}}
        \vspace{-1mm}
		\centerline{\small{(a) Input }}
	\end{minipage}
	\hfill
	\begin{minipage}{0.242\linewidth}
		\centerline{\includegraphics[width=1\linewidth]{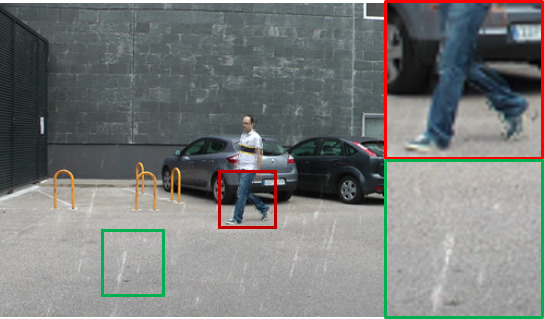}}
        \vspace{-1mm}
		\centerline{\small{(b) Garg et al. \cite{Garg2004Detection}}}
	\end{minipage}
	\hfill
	\begin{minipage}{.242\linewidth}
		\centerline{\includegraphics[width=1\linewidth]{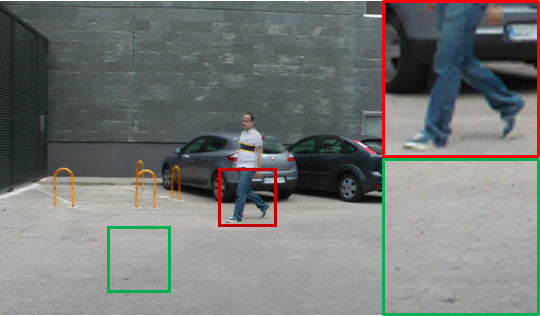}}
		\vspace{-1mm}
		\centerline{\small{(c)  Jiang et al. \cite{jiang2017cvpr}}}
	\end{minipage}
	\hfill
	\begin{minipage}{.242\linewidth}
		\centerline{\includegraphics[width=1\linewidth]{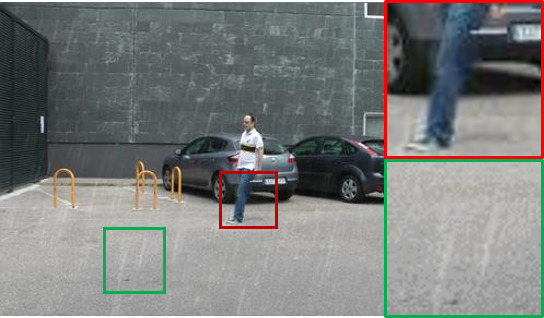}}
        \vspace{-1mm}
		\centerline{\small{(d) Ren et al. \cite{ren2017cvpr}}}
	\end{minipage}
    \vfill
	\begin{minipage}{0.242\linewidth}
		\centerline{\includegraphics[width=1\linewidth]{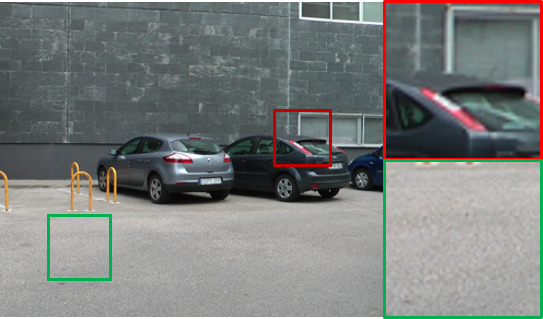}}
	\end{minipage}
	\hfill
     \begin{minipage}{0.242\linewidth}
	  \centerline{\includegraphics[width=1\linewidth]{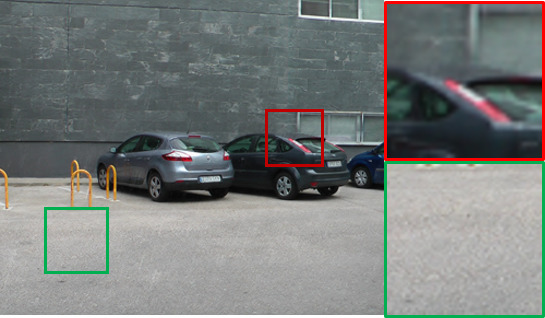}}
	 \end{minipage}
	\hfill
	\begin{minipage}{0.242\linewidth}
		\centerline{\includegraphics[width=1\linewidth]{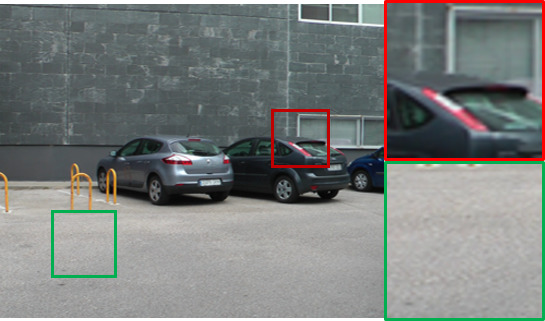}}
	\end{minipage}
	\hfill
	\begin{minipage}{0.242\linewidth}
		\centerline{\includegraphics[width=1\linewidth]{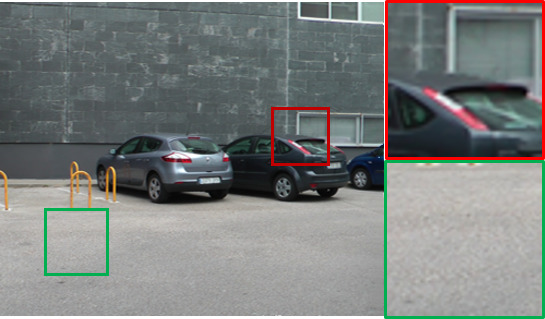}}
	\end{minipage}
    \vfill
	\begin{minipage}{0.242\linewidth}
		\centerline{\includegraphics[width=1\linewidth]{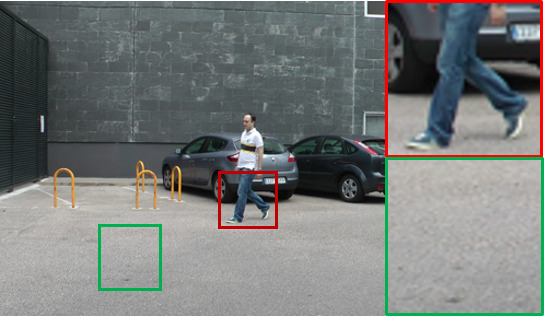}}
		\vspace{-1mm}
		\centerline{\small{(e) Groundtruth}}
	\end{minipage}
	\hfill
	\begin{minipage}{.242\linewidth}
	  \centerline{\includegraphics[width=1\linewidth]{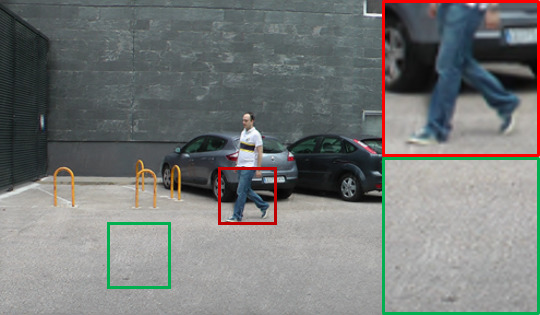}}
      \vspace{-1mm}
	  \centerline{\small{(f) Liu et al. \cite{Liu18Erase}}}
	\end{minipage}
	\hfill
	\begin{minipage}{.242\linewidth}
		\centerline{\includegraphics[width=1\linewidth]{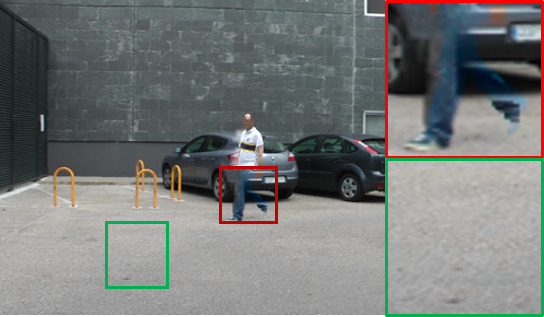}}
        \vspace{-1mm}
		\centerline{\small{(g) TMS-CSC}}
	\end{minipage}
	\hfill
	\begin{minipage}{0.242\linewidth}
		\centerline{\includegraphics[width=1\linewidth]{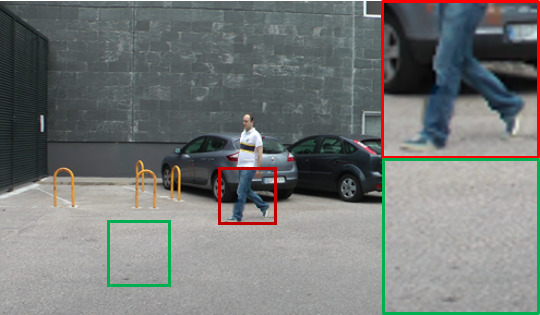}}
		\vspace{-1mm}
		\centerline{\small{(h) OTMS-CSC}}
	\end{minipage}
	\caption{
(a) Two typical input frame in a video with heavy rain. (b)-(h) Recovered frames obtained by different competing methods.
}
	\label{fig:human}
	\vspace{-5mm}
\end{figure*}

\begin{table*}[!htbp]\vspace{-2mm}
\caption{ Quantitative performance comparison of all competing methods on synthetic rainy and snowy videos.}\label{synmetrics}
\vspace{-3mm}
\makebox[\textwidth][c]{
\begin{tabular*}{0.78\paperwidth}{ @{} p{1.5cm}<{\centering}
|p{0.5cm}<{\centering}p{0.5cm}<{\centering}p{0.5cm}<{\centering}p{0.55cm}<{\centering}
|p{0.5cm}<{\centering}p{0.5cm}<{\centering}p{0.5cm}<{\centering}p{0.55cm}<{\centering}
|p{0.5cm}<{\centering}p{0.5cm}<{\centering}p{0.5cm}<{\centering}p{0.55cm}<{\centering}
|p{0.5cm}<{\centering}p{0.5cm}<{\centering}p{0.5cm}<{\centering}p{0.55cm}<{\centering} }
\Xhline{1.5pt}
Types &\multicolumn{12}{|c|}{Static videos}  &\multicolumn{4}{|c}{ Dynamic video}\\
\hline
Dataset &\multicolumn{4}{|c|}{Fig. \ref{fig:park} } &\multicolumn{4}{|c|}{Fig. \ref{fig:highway} } &\multicolumn{4}{|c}{Fig. \ref{fig:4080}} &\multicolumn{4}{|c}{Fig. \ref{fig:human} }\\
\hline
Metrics &PSNR &VIF &FSIM&SSIM    &PSNR &VIF &FSIM&SSIM     &PSNR &VIF &FSIM&SSIM  &PSNR &VIF &FSIM&SSIM\\
\Xhline{1.5pt}
Input                      &28.22 & 0.637  &0.935	 &0.927   & 23.82 &0.766  &0.970  &0.929 &27.93	&0.595	&0.859  &0.831 &29.32 &0.752 &0.995 &0.909 \\ 
Garg \cite{Garg2007Vision} &29.83 &	0.661  &0.955	 &0.946   & 24.64 &0.750  &0.972  &0.920 &35.87 &0.819	&0.957  &0.950 &36.11 &0.849 &0.977 &0.969 \\ 
Jiang\cite{jiang2017cvpr}  &31.01 &0.767   &0.967    &0.959   & 24.32 &0.713  &0.966  &0.929 &35.80	&0.779	&0.982  &0.977 &32.51 &0.693 &0.998 &0.960	\\ 
Ren  \cite{ren2017cvpr}    &28.26 &0.685   &0.970	 &0.962   &	23.52 &0.681  &0.966  &0.927 &30.34 &0.921	&0.753  &0.995 &31.33 &0.626 &0.994	&0.956 \\
Wei  \cite{wei2017should}  &29.76 &0.822   &0.991    &0.986	  & 24.43 &0.761  &0.973  &0.943 &34.58	&0.945	&0.996  &0.993 &- &- &- &- \\	
Liu  \cite{Liu18Erase}     &27.56 &0.626   &0.995    &0.941	  &22.19  &0.555  &0.946  &0.895 &31.56	&0.616	&0.996  &0.946 &34.69 &0.716 &0.998 &0.965\\
(T)MS-CSC &\textbf{33.89} &\textbf{0.865} &\textbf{0.992 }   &\textbf{0.992}  &25.37 &0.790 &\textbf{0.980}  &\textbf{0.957} &42.95	&0.980	&\textbf{0.999} &0.997  &36.90	&0.862	&\textbf{0.999 } &\textbf{0.982}\\
OTMS-CSC                     &32.58 & 0.853  &0.991 &0.989 &\textbf{25.91} &\textbf{0.796} &0.979 &\textbf{0.957} &\textbf{46.29} &\textbf{0.988}	&\textbf{0.999} &\textbf{0.999} &\textbf{37.65}	&\textbf{0.869}	&0.983 &0.966 \\	

\Xhline{1.5pt}
\end{tabular*}
}\vspace{-2mm}
\end{table*}

\section{Experimental Results}
In this section, we evaluate the performance of our method on videos with synthetic and real rain/snow in both quantitative and qualitative perspectives. Some state-of-the-art video rain/snow removal methods have also been implemented for comparison, including Garg et al.~\cite{Garg2004Detection}\footnote{http://www.cs.columbia.edu/CAVE/projects/camera rain/}, Jiang et al.~\cite{jiang2017cvpr}\footnote{Code is provided by the authors}, Ren et al.~\cite{ren2017cvpr}\footnote{http://vision.sia.cn/our\%20team/RenWeihong-homepage/vision-renweihong\%28English\%29.html}, Wei et al.~\cite{wei2017should}\footnote{http://vision.sia.cn/our\%20team/RenWeihong-homepage/vision-renweihong\%28English\%29.html}, Li et al.~\cite{Li18Multiscale} and Liu et al.~\cite{Liu18Erase}\footnote{https://github.com/flyywh/J4RNet-Deep-Video-Deraining-CVPR-2018}.
Note that these methods contain both model-driven MAP-based and data-driven DL representative state-of-the-arts for a comprehensive comparison. Furthermore, through introducing traditional offline alignment strategy into the MS-CSC model, called transformed MS-CSC or TMS-CSC, this offline method can also be ameliorated to adapt to videos with background transformations.
All experiments were implemented on a PC with i7 CPU and 32G RAM. To make a sufficiently comprehensive comparison, more video demonstrations on the obtained results by completing methods have been reported in our specifically constructed website\footnote{https://sites.google.com/view/onlinetmscsc/ \label{footnote1}} for easy observation.


\subsection{Experiments on Videos with Synthetic Rain/Snow}
We first introduce experiments executed on four videos with synthetic rain/snow, three with static backgrounds, as shown in Fig. \ref{fig:park} -- \ref{fig:4080}, and one with evidently dynamic background with evident translations among adjacent frames, as depicted in Fig. \ref{fig:human}. The clean videos as pshown in Fig.~\ref{fig:highway} and Fig.\ref{fig:human} are downloaded from CDNET database\cite{goyette2012changedetection}\footnote{http://www.changedetection.net}, and those of Fig. \ref{fig:park} and Fig. \ref{fig:4080} are downloaded from Youtube\footnote{https://www.youtube.com/watch?v=aOhdnllS0\_k} and Xi'an Jiaotong University surveillance camera, respectively. Especially, the videos as shown in \ref{fig:highway} and \ref{fig:4080} contain heavy rain and snow forming serious occlusions to background scene and foreground objects throughout the video sequences, respectively. The rain/snow with various types were synthetically generated by Photoshop on a black background.

\begin{figure*}[!htb]
	\begin{minipage}{0.242\linewidth}
		\centerline{\includegraphics[width=1\linewidth]{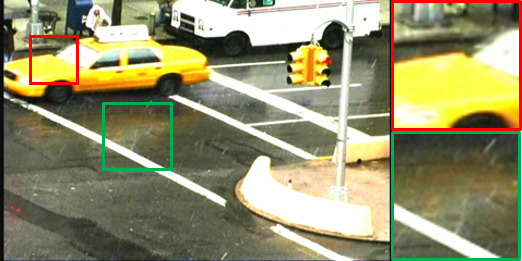}}
		\vspace{-1mm}
		\centerline{\small{(a) Input}}
	\end{minipage}
	\hfill
	\begin{minipage}{.242\linewidth}
		\centerline{\includegraphics[width=1\linewidth]{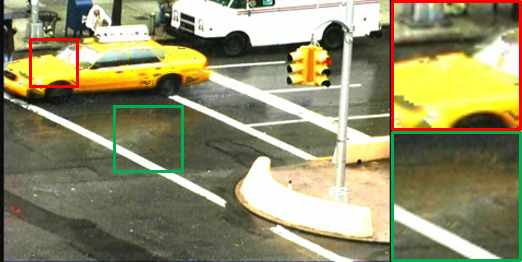}}
		\vspace{-1mm}
		\centerline{\small{(b)  Garg et al. \cite{Garg2007Vision}}}
	\end{minipage}
	\hfill
	\begin{minipage}{.242\linewidth}
		\centerline{\includegraphics[width=1\linewidth]{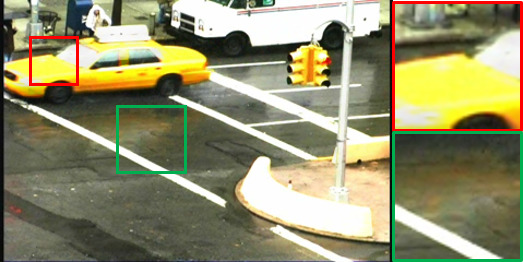}}
		\vspace{-1mm}
		\centerline{\small{(c)  Jiang et al. \cite{jiang2017cvpr}}}
	\end{minipage}
	\hfill
	\begin{minipage}{0.242\linewidth}
		\centerline{\includegraphics[width=1\linewidth]{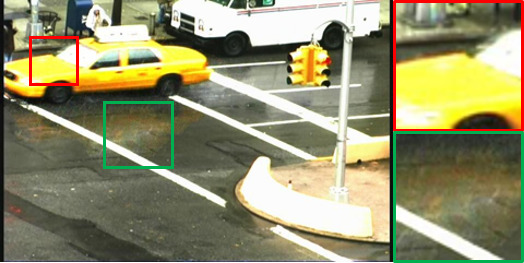}}
		\vspace{-1mm}
		\centerline{\small{(d) Ren et al. \cite{ren2017cvpr}}}
	\end{minipage}
	\hfill
	\begin{minipage}{0.242\linewidth}
		\centerline{\includegraphics[width=1\linewidth]{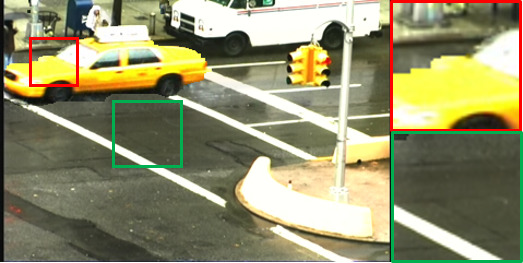}}
		\vspace{-1mm}
		\centerline{\small{(e) Wei et al. \cite{wei2017should}}}
	\end{minipage}
	\hfill
	\begin{minipage}{0.242\linewidth}
		\centerline{\includegraphics[width=1\linewidth]{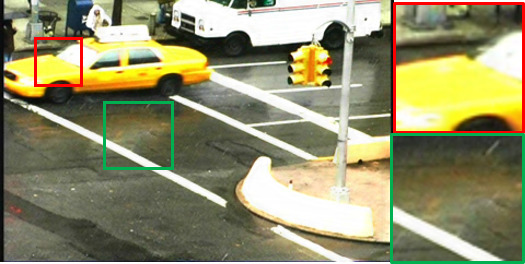}}
		\vspace{-1mm}
		\centerline{\small{(f) Liu et al. \cite{Liu18Erase}}}
	\end{minipage}
	\hfill
	\begin{minipage}{0.242\linewidth}
		\centerline{\includegraphics[width=1\linewidth]{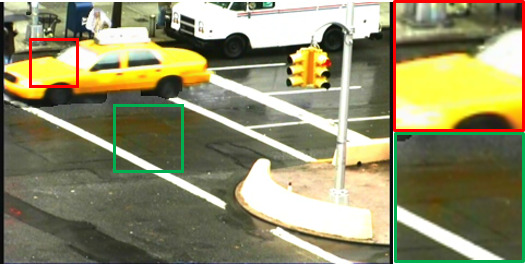}}
		\vspace{-1mm}
		\centerline{\small{(g) MS-CSC}}
	\end{minipage}
	\hfill
	\begin{minipage}{0.242\linewidth}
		\centerline{\includegraphics[width=1\linewidth]{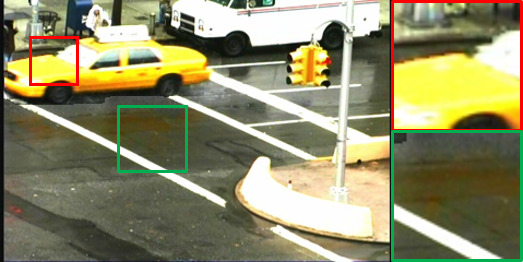}}
		\vspace{-1mm}
		\centerline{\small{(h) OTMS-CSC}}
	\end{minipage}
	\caption{ (a) An input frame of a real rainy video with complex moving objects. (b)-(h) Recovered frames obtained by different competing methods.
 }
	\label{fig:compfinal}
	\vspace{-2mm}
\end{figure*}

\begin{figure*}[!htb]
	\begin{minipage}{0.242\linewidth}
		\centerline{\includegraphics[width=1\linewidth]{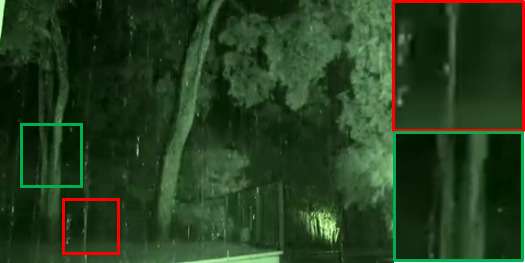}}
		\vspace{-1mm}
		\centerline{\small{(a) Input}}
	\end{minipage}
	\hfill
	\begin{minipage}{.242\linewidth}
		\centerline{\includegraphics[width=1\linewidth]{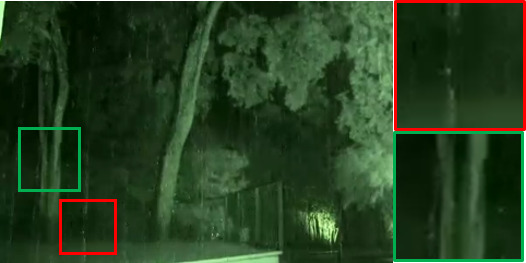}}
		\vspace{-1mm}
		\centerline{\small{(b)  Garg et al.~\cite{Garg2007Vision}}}
	\end{minipage}
	\hfill
	\begin{minipage}{.242\linewidth}
		\centerline{\includegraphics[width=1\linewidth]{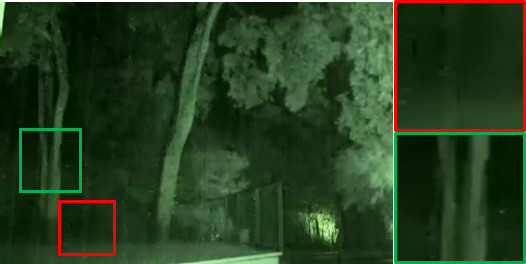}}
		\vspace{-1mm}
		\centerline{\small{(c)  Jiang et al.~\cite{jiang2017cvpr}}}
	\end{minipage}
	\hfill
	\begin{minipage}{0.242\linewidth}
		\centerline{\includegraphics[width=1\linewidth]{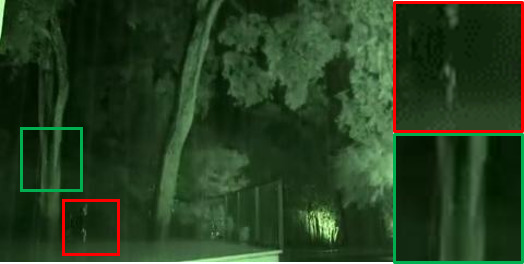}}
		\vspace{-1mm}
		\centerline{\small{(d) Ren et al.~\cite{ren2017cvpr}}}
	\end{minipage}
	\hfill
	\begin{minipage}{0.242\linewidth}
		\centerline{\includegraphics[width=1\linewidth]{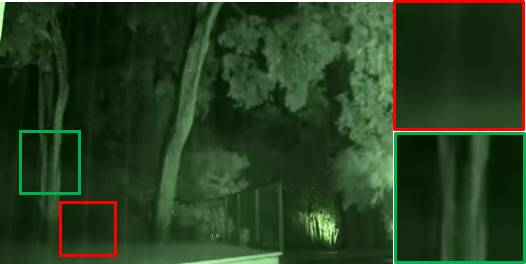}}
		\vspace{-1mm}
		\centerline{\small{(e) Wei et al.~\cite{wei2017should}}}
	\end{minipage}
	\hfill
	\begin{minipage}{0.242\linewidth}
		\centerline{\includegraphics[width=1\linewidth]{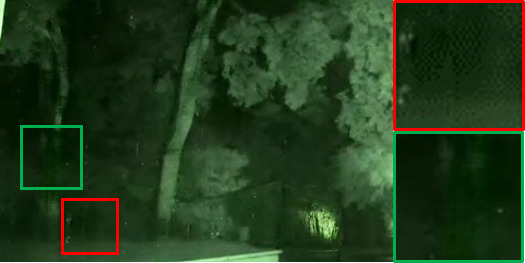}}
		\vspace{-1mm}
		\centerline{\small{(f) Liu et al.~\cite{Liu18Erase}}}
	\end{minipage}
	\hfill
	\begin{minipage}{0.242\linewidth}
		\centerline{\includegraphics[width=1\linewidth]{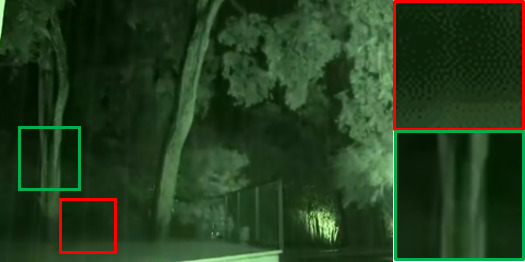}}
		\vspace{-1mm}
		\centerline{\small{(g) MS-CSC}}
	\end{minipage}
	\hfill
	\begin{minipage}{0.242\linewidth}
		\centerline{\includegraphics[width=1\linewidth]{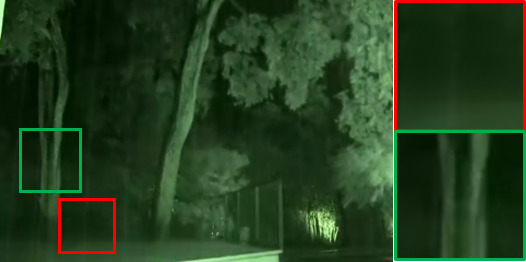}}
		\vspace{-1mm}
		\centerline{\small{(h) OTMS-CSC}}
	\end{minipage}
	\caption{
 (a) An input frame of a real rainy video captured at night. (b)-(h) Recovered frames obtained by different competing methods.
}
	\label{fig:night}
	\vspace{-2mm}
\end{figure*}

From Fig.~\ref{fig:park} and Fig.~\ref{fig:highway}, we can easily observe that the compared methods proposed by Garg et al., Jiang et al. and Liu et al. haven't completely removed the rain streaks and the method proposed by Ren et al. has not finely kept the shape of the moving object when removing the rain streaks. Besides, as shown in the second row of Fig.~\ref{fig:park}, the rain layer extracted by all other competing methods contain more or less additional background information. Comparatively, the proposed OTMS-CSC method, as well as its offline version MS-CSC, can finely remove the rain in the video and well maintain the shape and texture details.

From Fig.~\ref{fig:4080}, it can be seen that most competing methods have not finely removed snow from the video, and the separated snow layer by MS-CSC method improperly contains certain moving objects. Comparatively, the OTMS-CSC method has a better performance in both snow removing and background/foreground detail preservation.

For dynamic videos as shown in Fig.~\ref{fig:human}, we can observe that the methods proposed by Garg et al. and Ren et al. have not fully removed the rain details on the images. The method proposed by Ren et al. as well as the offline TMS-CSC method have not finely preserved the structure of the moving objects from foreground, and that proposed by Jiang et al. has also not done well in background detail preservation (like the texture of wall). Comparatively, our proposed OTMS-CSC method attains a relatively better performance in both aspects.

Quantitative comparisons are also presented in Table~\ref{synmetrics}, which fully complies with the aforementioned visual observations. Specifically, we adopt four image quality assessment (IQA) metrics to evaluate the performance of all competing methods, namely, PSNR, VIF~\cite{sheikh2006image}, FSIM~\cite{zhang2011fsim} and SSIM~\cite{wang2004image}. From the table, it can be seen that our proposed OTMS-CSC model can perform best or the second best in almost all cases in terms of all IQAs, as compared with other competing methods. Considering that all other methods are implemented on the entire video (iteratively utilizing the video multiple times) or need additionally pre-collected training data while our method is sequentially implemented in the video sequence (i.e., each frame is only iterated one time and then dropped out), it should be rational to say our method is efficient.

\begin{figure*}[!htb]
	\begin{minipage}{0.242\linewidth}
		\centerline{\includegraphics[width=1\linewidth]{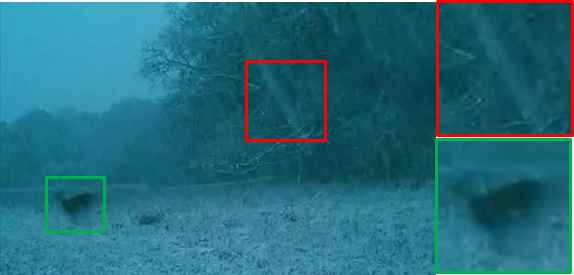}}
		\vspace{-1mm}
		\centerline{\small{(a) Input}}
	\end{minipage}
	\hfill
	\begin{minipage}{.242\linewidth}
		\centerline{\includegraphics[width=1\linewidth]{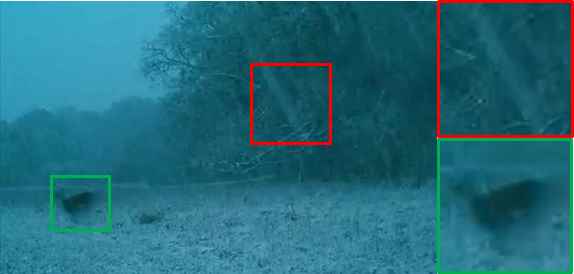}}
		\vspace{-1mm}
		\centerline{\small{(b)  Garg et al. \cite{Garg2007Vision}}}
	\end{minipage}
	\hfill
	\begin{minipage}{.242\linewidth}
		\centerline{\includegraphics[width=1\linewidth]{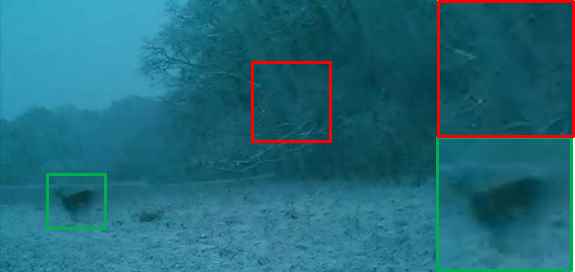}}
		\vspace{-1mm}
		\centerline{\small{(c)  Jiang et al. \cite{jiang2017cvpr}}}
	\end{minipage}
	\hfill
	\begin{minipage}{0.242\linewidth}
		\centerline{\includegraphics[width=1\linewidth]{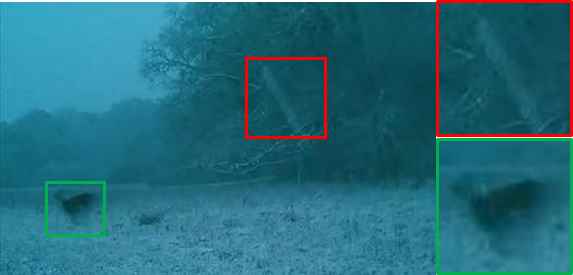}}
		\vspace{-1mm}
		\centerline{\small{(d) Ren et al. \cite{ren2017cvpr}}}
	\end{minipage}
	\hfill
	\begin{minipage}{0.242\linewidth}
		\centerline{\includegraphics[width=1\linewidth]{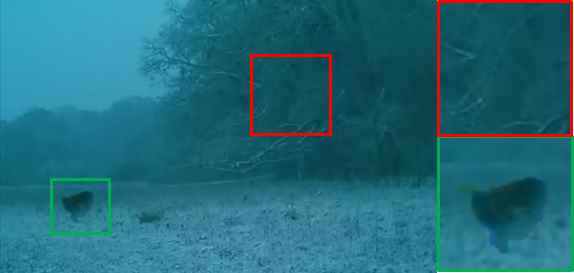}}
		\vspace{-1mm}
		\centerline{\small{(e) Wei et al. \cite{wei2017should}}}
	\end{minipage}
	\hfill
	\begin{minipage}{0.242\linewidth}
		\centerline{\includegraphics[width=1\linewidth]{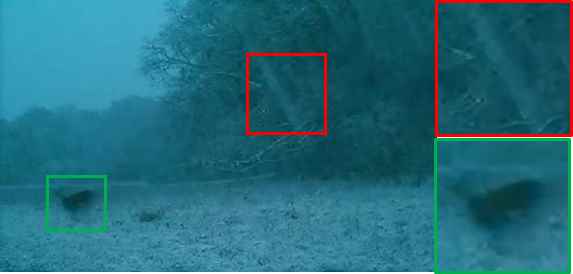}}
		\vspace{-1mm}
		\centerline{\small{(f) Liu et al. \cite{Liu18Erase}}}
	\end{minipage}
	\hfill
	\begin{minipage}{0.242\linewidth}
		\centerline{\includegraphics[width=1\linewidth]{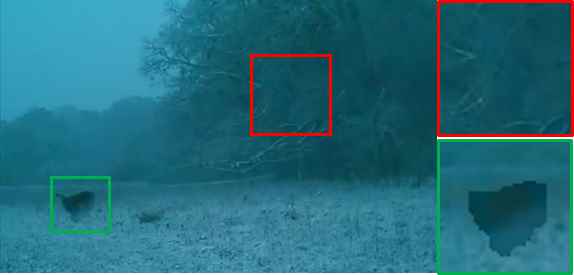}}
		\vspace{-1mm}
		\centerline{\small{(g) MS-CSC}}
	\end{minipage}
	\hfill
	\begin{minipage}{0.242\linewidth}
		\centerline{\includegraphics[width=1\linewidth]{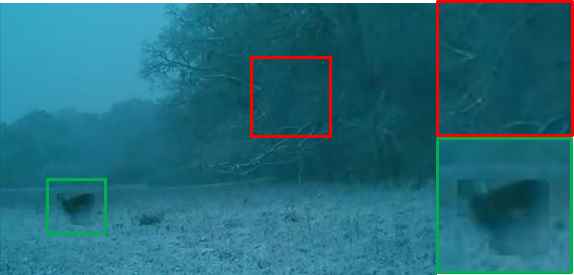}}
		\vspace{-1mm}
		\centerline{\small{(h) OTMS-CSC}}
	\end{minipage}
	\caption{
(a) An input frame of a real snowy video with poor visibility. (b)-(h) Recovered frames obtained by different competing methods.
}
	\label{fig:animal}
	\vspace{-2mm}
\end{figure*}

\begin{figure*}[!htb]
	\begin{minipage}{0.242\linewidth}
		\centerline{\includegraphics[width=1\linewidth]{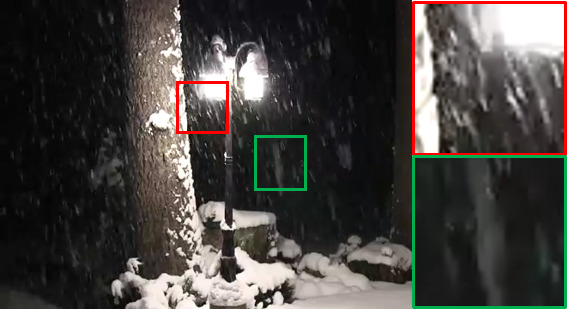}}
        \vspace{-1mm}
		\centerline{\small{(a) Input}}
	\end{minipage}
	\hfill
	\begin{minipage}{.242\linewidth}
		\centerline{\includegraphics[width=1\linewidth]{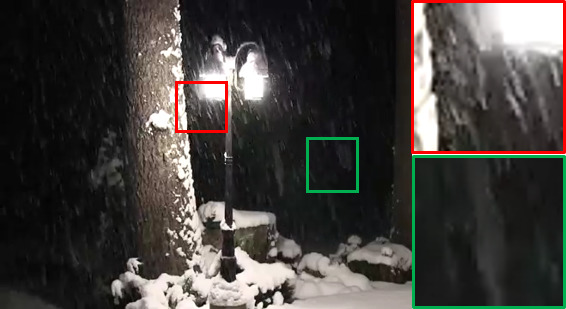}}
        \vspace{-1mm}
		\centerline{\small{(b) Garg et al. \cite{Garg2007Vision}}}
	\end{minipage}
	\hfill
	\begin{minipage}{.242\linewidth}
		\centerline{\includegraphics[width=1\linewidth]{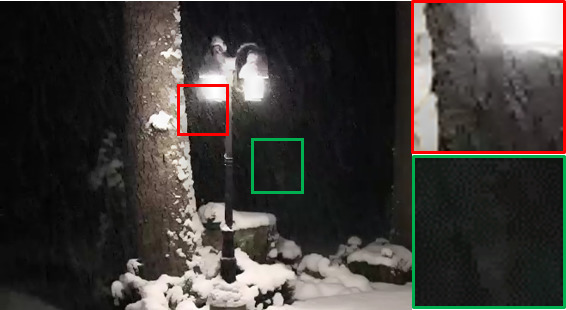}}
        \vspace{-1mm}
		\centerline{\small{(c) Jiang et al. \cite{jiang2017cvpr}}}
	\end{minipage}
	\hfill
	\begin{minipage}{0.242\linewidth}
		\centerline{\includegraphics[width=1\linewidth]{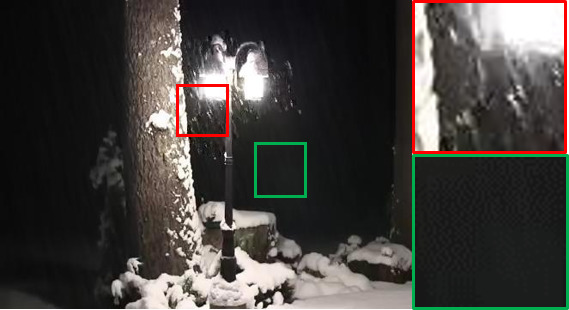}}
        \vspace{-1mm}
		\centerline{\small{(d) Ren et al. \cite{ren2017cvpr}}}
	\end{minipage}
	\hfill
	\begin{minipage}{0.242\linewidth}
		\centerline{\includegraphics[width=1\linewidth]{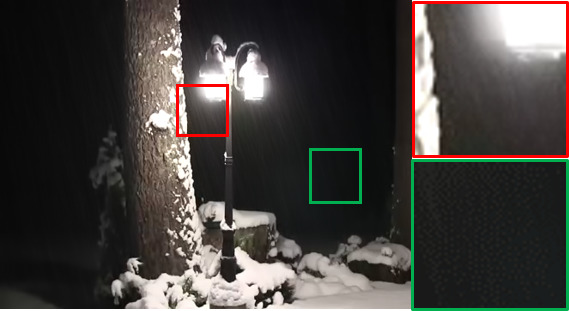}}
        \vspace{-1mm}
		\centerline{\small{(e) Wei et al. \cite{wei2017should}}}
	\end{minipage}
	\hfill
	\begin{minipage}{0.242\linewidth}
		\centerline{\includegraphics[width=1\linewidth]{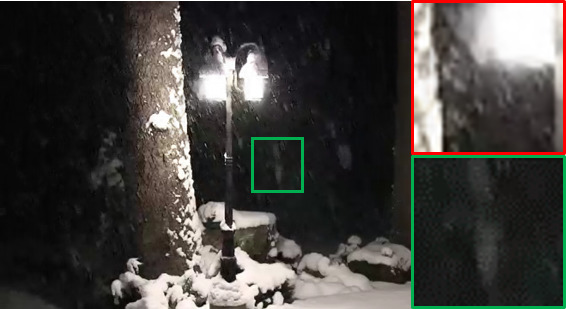}}
        \vspace{-1mm}
		\centerline{\small{(f) Liu et al. \cite{Liu18Erase}}}
	\end{minipage}
	\hfill
	\begin{minipage}{0.242\linewidth}
		\centerline{\includegraphics[width=1\linewidth]{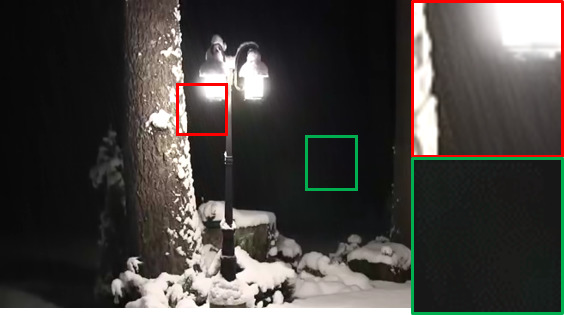}}
        \vspace{-1mm}
		\centerline{\small{(g) MS-CSC}}
	\end{minipage}
	\hfill
	\begin{minipage}{0.242\linewidth}
		\centerline{\includegraphics[width=1\linewidth]{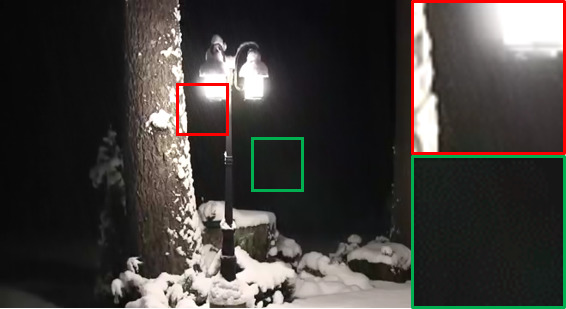}}
        \vspace{-1mm}
		\centerline{\small{(h) OTMS-CSC}}
	\end{minipage}
	\caption{
(a) An input frame of a real video with dynamic snow shapts. (b)-(h) Recovered frames obtained by different competing methods.}
	\label{fig:light}
	\vspace{-1mm}
\end{figure*}

\begin{figure*}[!htb]\vspace{-2mm}
\begin{minipage}{0.242\linewidth}
  \centerline{\includegraphics[width=1\linewidth]{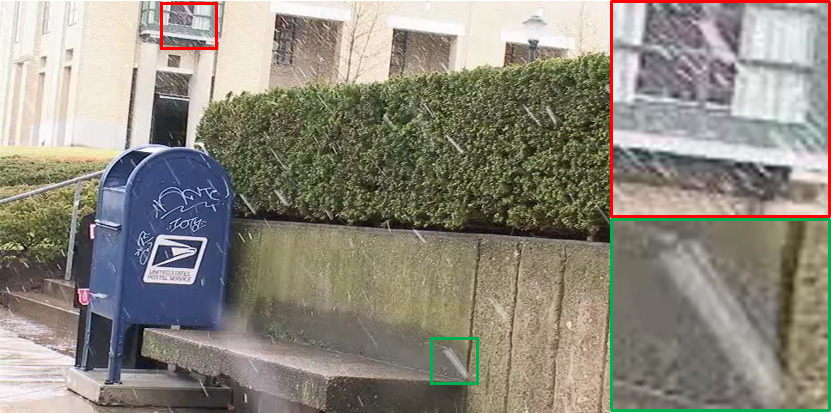}}
\end{minipage}
\hfill
\begin{minipage}{0.242\linewidth}
  \centerline{\includegraphics[width=1\linewidth]{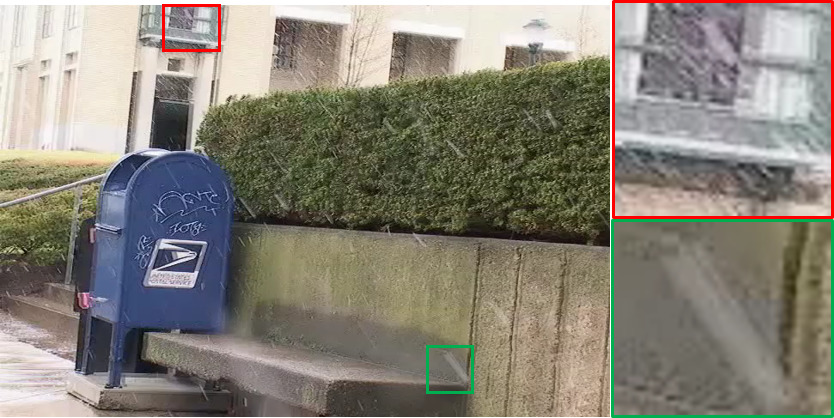}}
\end{minipage}
\hfill
\begin{minipage}{.242\linewidth}
  \centerline{\includegraphics[width=1\linewidth]{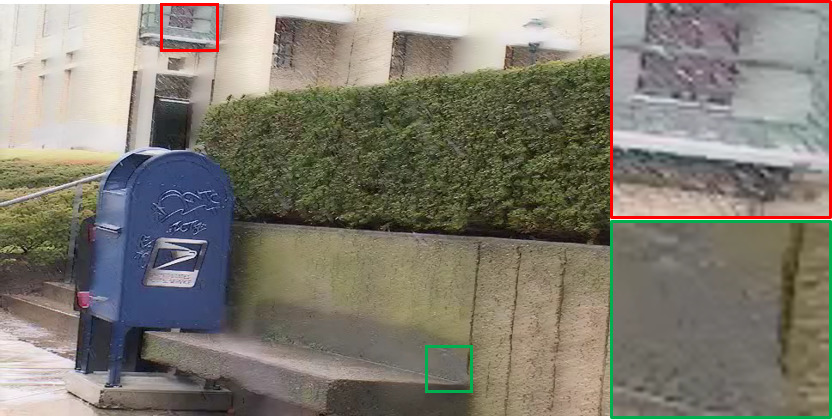}}
\end{minipage}
\hfill
\begin{minipage}{0.242\linewidth}
  \centerline{\includegraphics[width=1\linewidth]{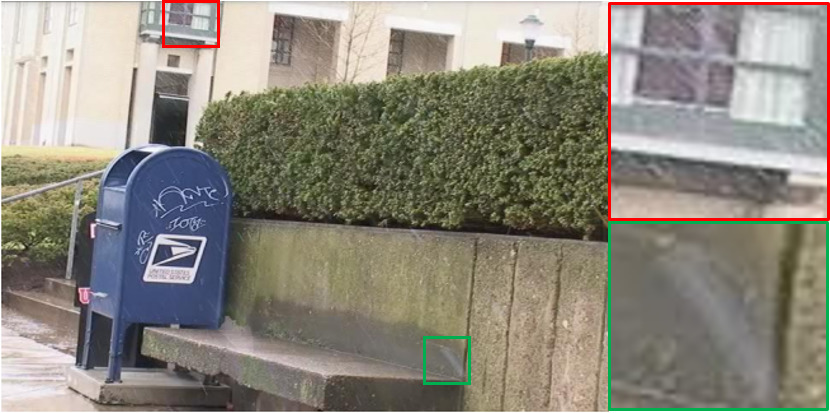}}
\end{minipage}
\vfill
\begin{minipage}{.242\linewidth}
  \centerline{\includegraphics[width=1\linewidth]{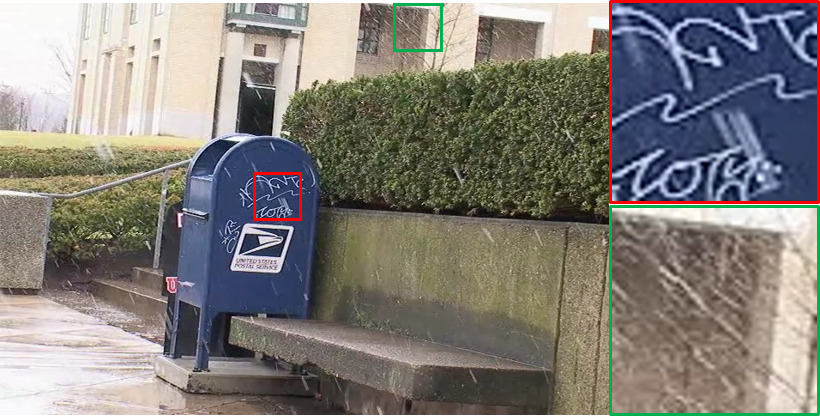}}
  \vspace{-1mm}
  \centerline{\small{(a) Input }}
\end{minipage}
\hfill
\begin{minipage}{0.242\linewidth}
  \centerline{\includegraphics[width=1\linewidth]{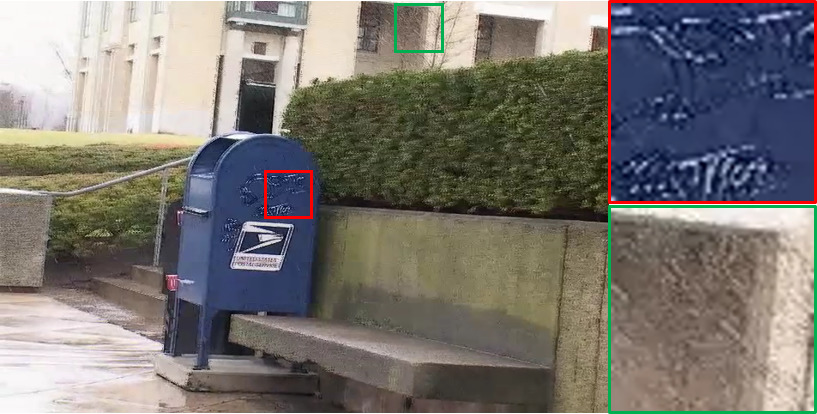}}
  \vspace{-1mm}
  \centerline{\small{(b) Garg et al. \cite{Garg2004Detection}}}
\end{minipage}
\hfill
\begin{minipage}{.242\linewidth}
  \centerline{\includegraphics[width=1\linewidth]{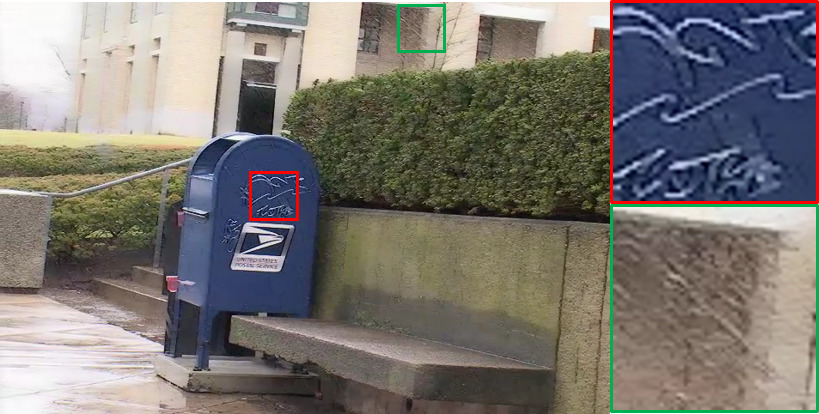}}
  \vspace{-1mm}
  \centerline{\small{(c)  Jiang et al. \cite{jiang2017cvpr}}}
\end{minipage}
\hfill
\begin{minipage}{.242\linewidth}
  \centerline{\includegraphics[width=1\linewidth]{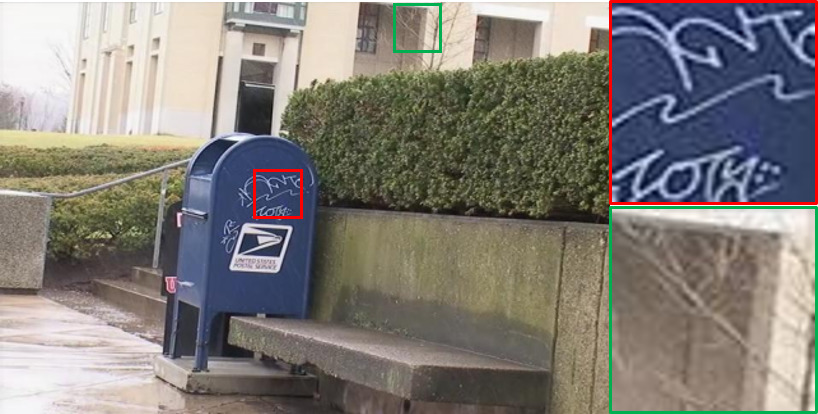}}
  \vspace{-1mm}
  \centerline{\footnotesize{(d) Ren et al. \cite{ren2017cvpr}}}
\end{minipage}
\vfill
\begin{minipage}{0.242\linewidth}
  \centerline{\includegraphics[width=1\linewidth]{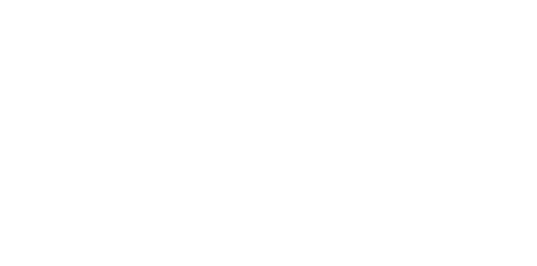}}
\end{minipage}
\hfill
\begin{minipage}{0.242\linewidth}
  \centerline{\includegraphics[width=1\linewidth]{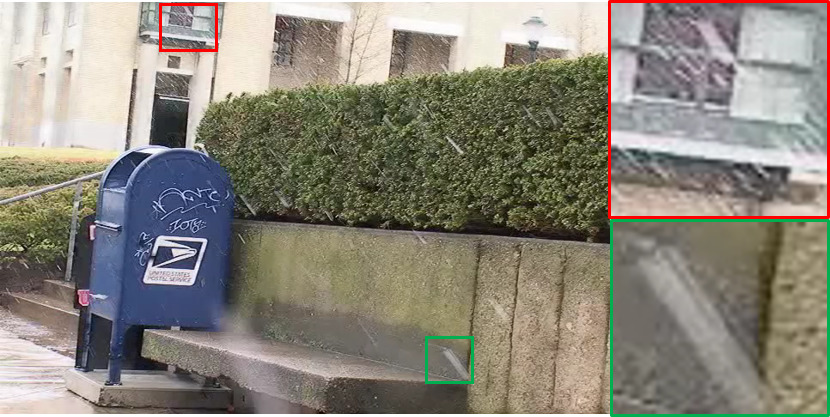}}
\end{minipage}
\hfill
\begin{minipage}{0.242\linewidth}
  \centerline{\includegraphics[width=1\linewidth]{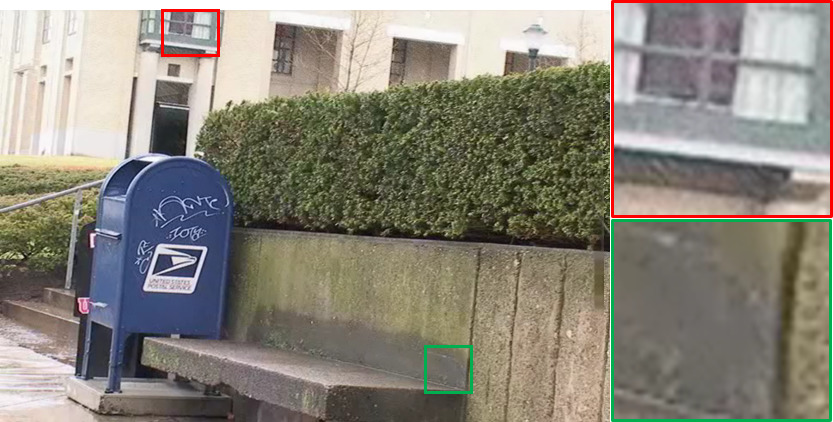}}
\end{minipage}
\hfill
\begin{minipage}{0.242\linewidth}
  \centerline{\includegraphics[width=1\linewidth]{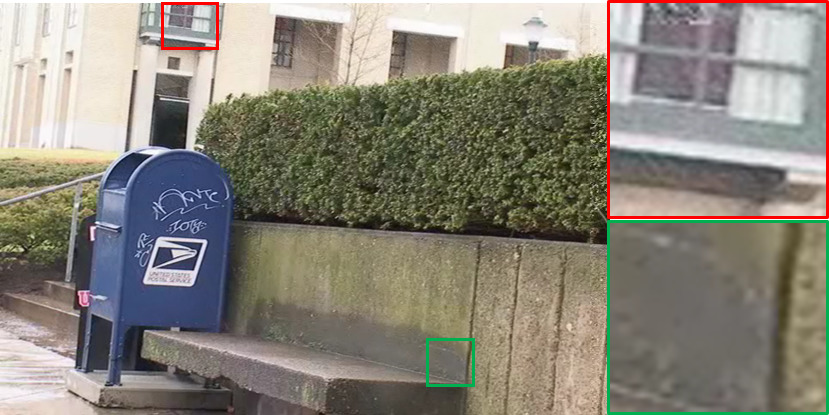}}
\end{minipage}
\vfill
\begin{minipage}{0.242\linewidth}
  \centerline{\includegraphics[width=1\linewidth]{postbox//8}}
\end{minipage}
\hfill
\begin{minipage}{.242\linewidth}
  \centerline{\includegraphics[width=1\linewidth]{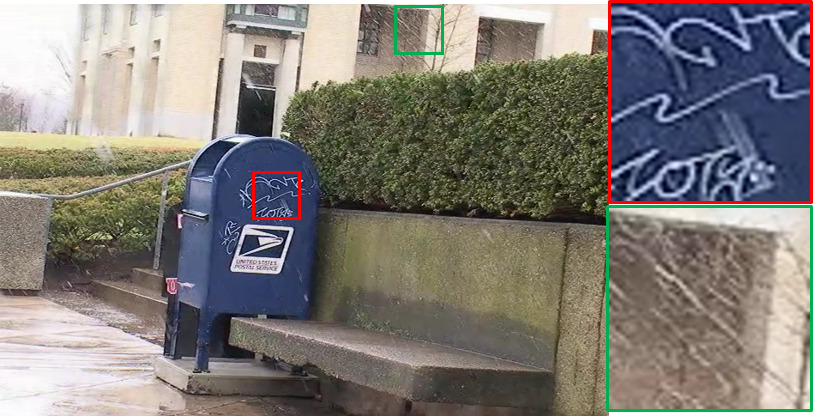}}
  \vspace{-1mm}
  \centerline{\small{(e) Liu et al. \cite{Liu18Erase}}}
\end{minipage}
\hfill
\begin{minipage}{.242\linewidth}
  \centerline{\includegraphics[width=1\linewidth]{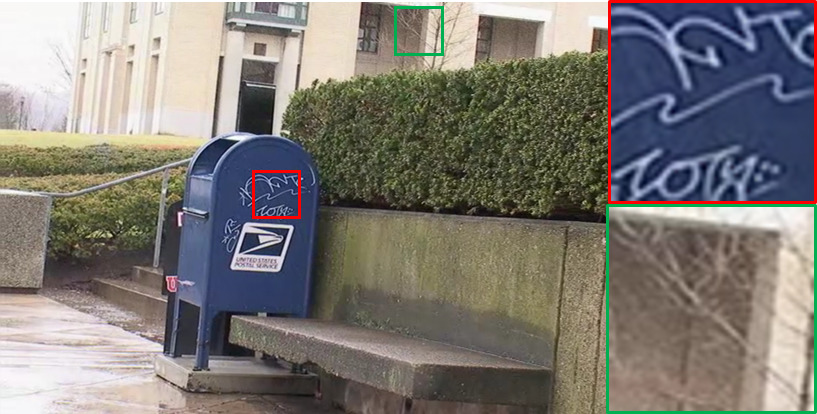}}
  \vspace{-1mm}
  \centerline{\small{(f) TMS-CSC}}
\end{minipage}
\hfill
\begin{minipage}{0.242\linewidth}
  \centerline{\includegraphics[width=1\linewidth]{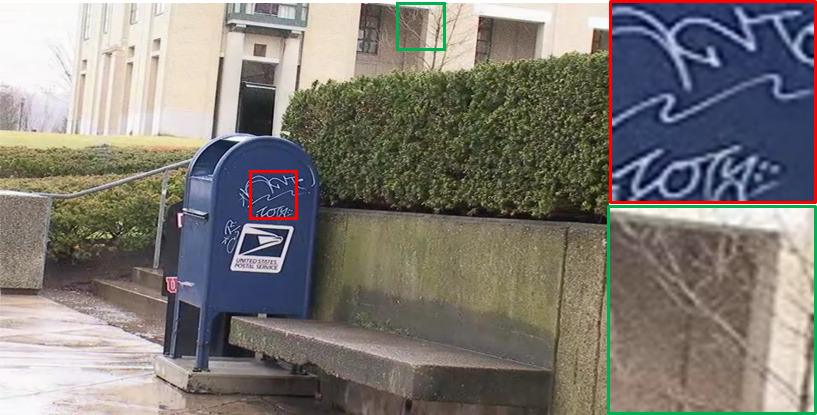}}
  \vspace{-1mm}
  \centerline{\small{(g) OTMS-CSC}}
\end{minipage}
\caption{(a) Two input frame of a real snowy video with fast horizontal movement. (b)-(h) Recovered frames obtained by different competing methods.
}
\label{fig:postbox}
\vspace{-1mm}
\end{figure*}

\begin{figure*}[!htb]
\begin{minipage}{0.242\linewidth}
  \centerline{\includegraphics[width=1\linewidth]{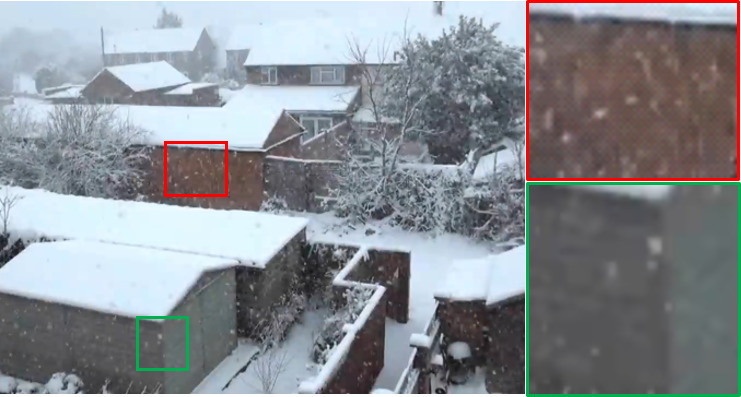}}
\end{minipage}
\hfill
\begin{minipage}{0.242\linewidth}
  \centerline{\includegraphics[width=1\linewidth]{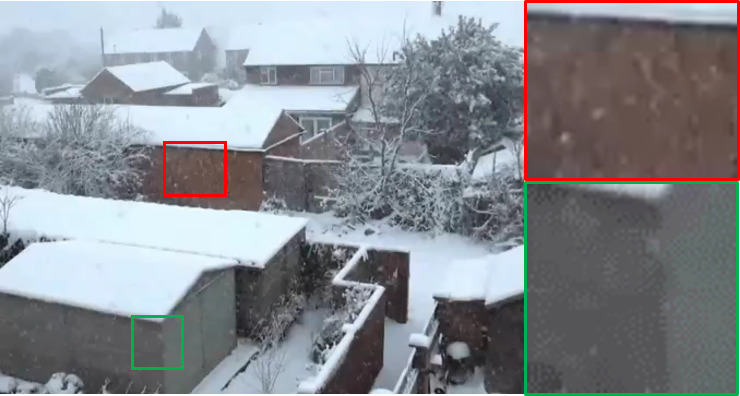}}
\end{minipage}
\hfill
\begin{minipage}{.242\linewidth}
  \centerline{\includegraphics[width=1\linewidth]{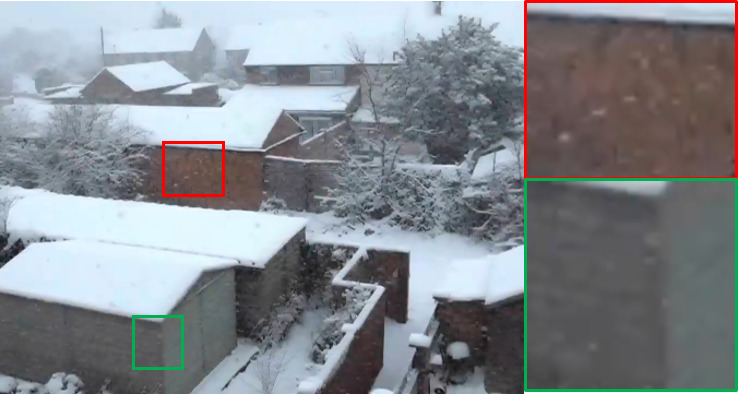}}
\end{minipage}
\hfill
\begin{minipage}{0.242\linewidth}
  \centerline{\includegraphics[width=1\linewidth]{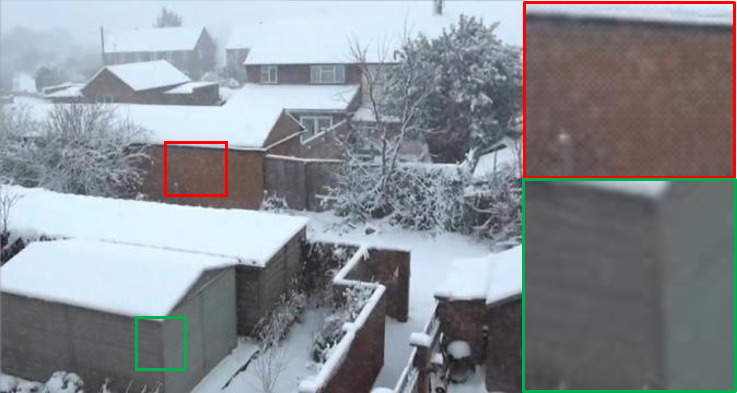}}
\end{minipage}
\vfill
\begin{minipage}{0.242\linewidth}
  \centerline{\includegraphics[width=1\linewidth]{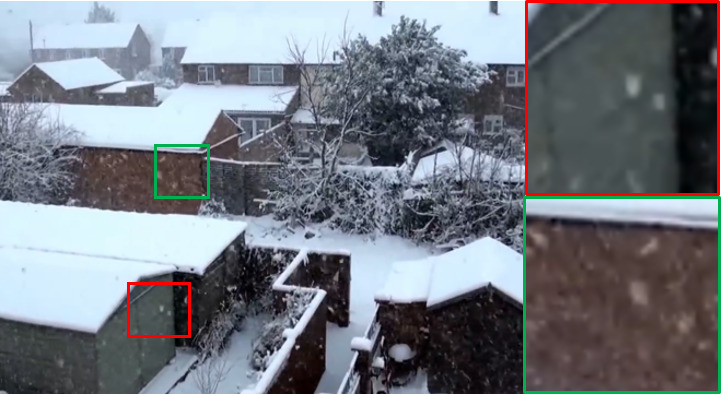}}
  \vspace{-1mm}
  \centerline{\small{(a) Input}}
\end{minipage}
\hfill
\begin{minipage}{0.242\linewidth}
  \centerline{\includegraphics[width=1\linewidth]{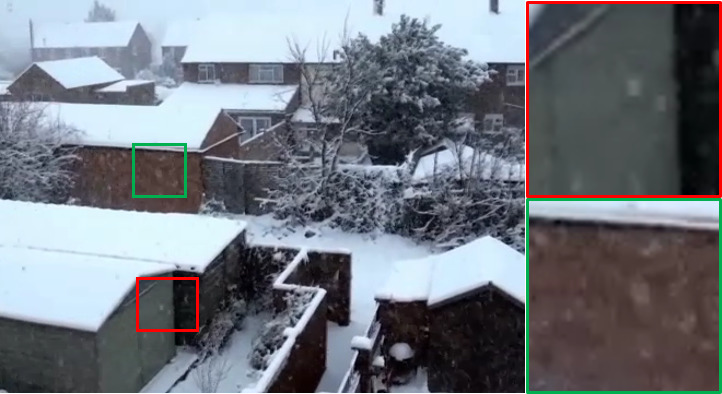}}
  \vspace{-1mm}
  \centerline{\small{(b) Garg et al. \cite{Garg2004Detection} }}
\end{minipage}
\hfill
\begin{minipage}{0.242\linewidth}
  \centerline{\includegraphics[width=1\linewidth]{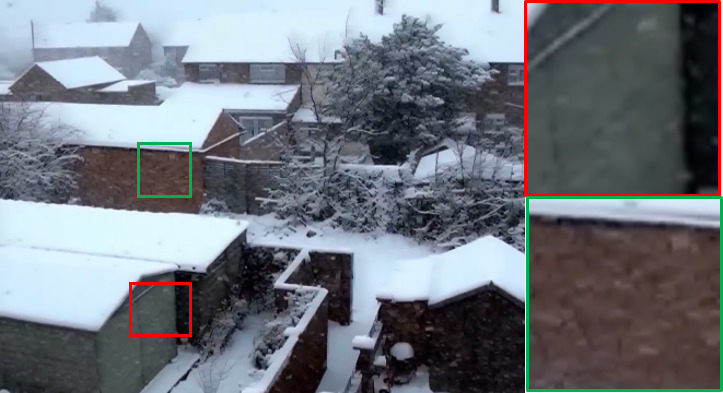}}
  \vspace{-1mm}
  \centerline{\small{(c) Jiang et al. \cite{jiang2017cvpr}}}
\end{minipage}
\hfill
\begin{minipage}{0.242\linewidth}
  \centerline{\includegraphics[width=1\linewidth]{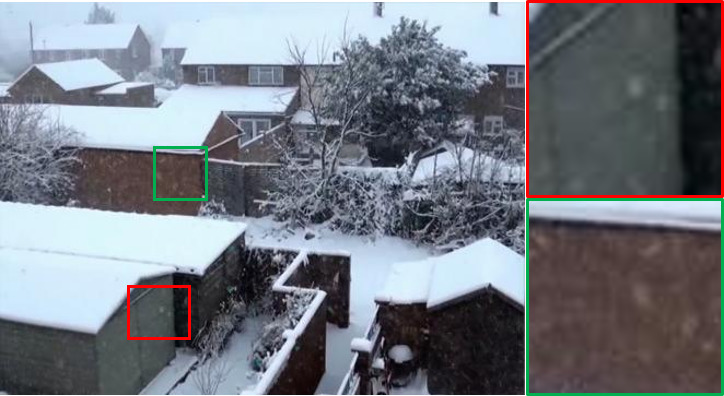}}
  \vspace{-1mm}
  \centerline{\small{(d) Ren et al. \cite{ren2017cvpr}}}
\end{minipage}
\vfill
\begin{minipage}{0.242\linewidth}
  \centerline{\includegraphics[width=1\linewidth]{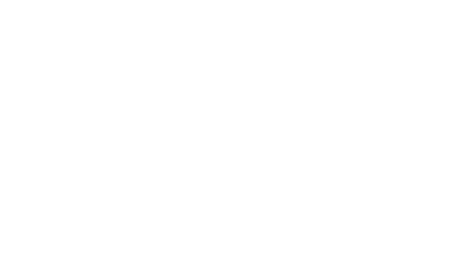}}
\end{minipage}
\hfill
\begin{minipage}{0.242\linewidth}
  \centerline{\includegraphics[width=1\linewidth]{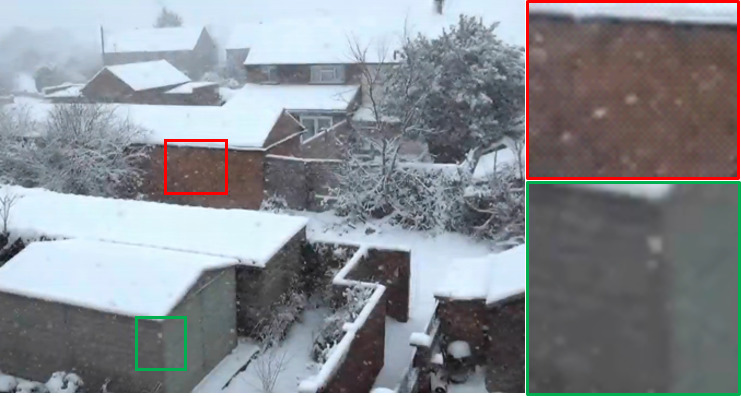}}
\end{minipage}
\hfill
\begin{minipage}{0.242\linewidth}
  \centerline{\includegraphics[width=1\linewidth]{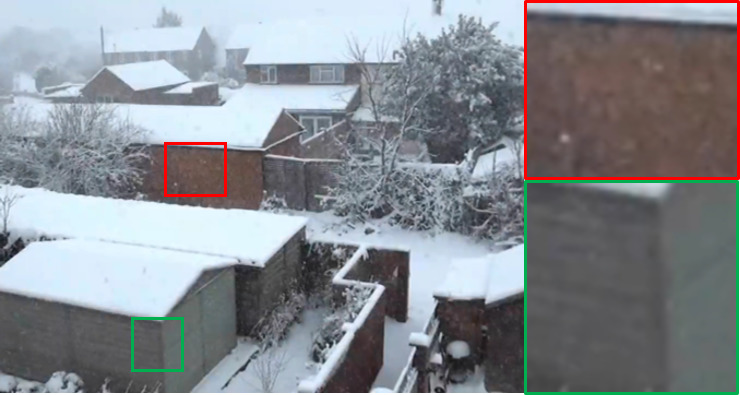}}
\end{minipage}
\hfill
\begin{minipage}{0.242\linewidth}
  \centerline{\includegraphics[width=1\linewidth]{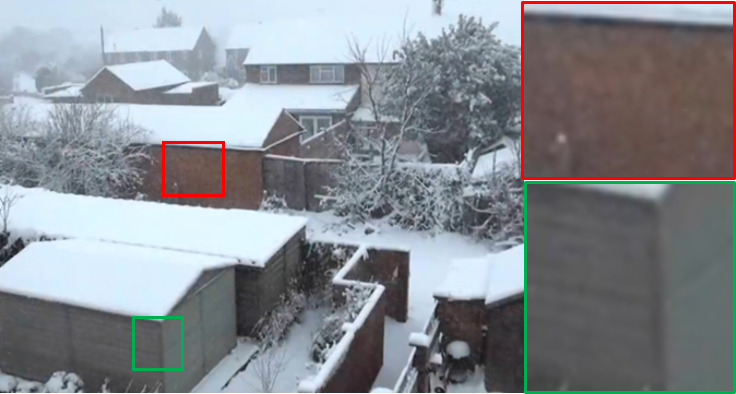}}
\end{minipage}
\vfill
\begin{minipage}{0.242\linewidth}
  \centerline{\includegraphics[width=1\linewidth]{t_london//9}}
\end{minipage}
\hfill
\begin{minipage}{0.242\linewidth}
  \centerline{\includegraphics[width=1\linewidth]{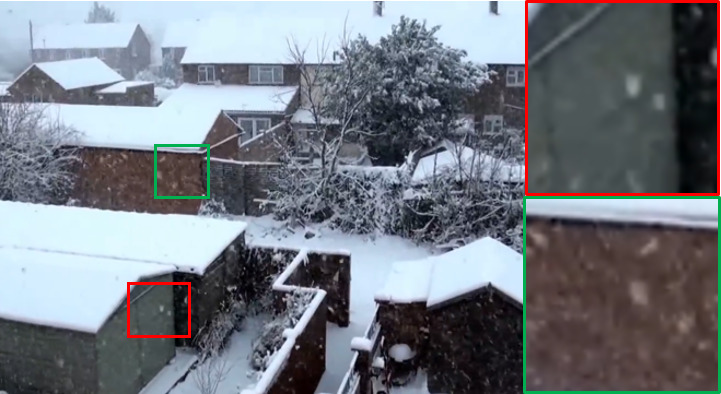}}
  \vspace{-1mm}
  \centerline{\small{(e) Liu et al. \cite{Liu18Erase}}}
\end{minipage}
\hfill
\begin{minipage}{0.242\linewidth}
  \centerline{\includegraphics[width=1\linewidth]{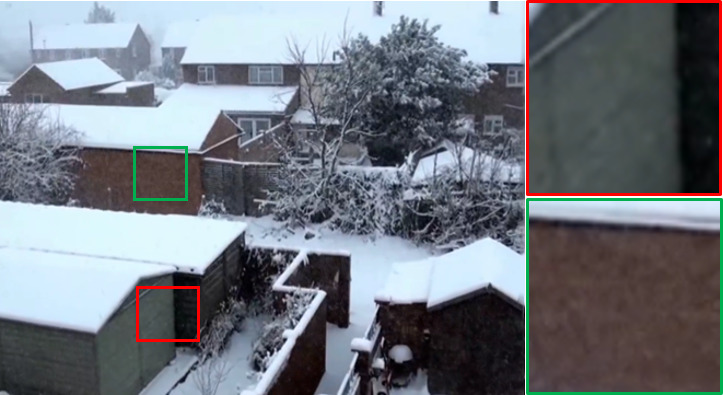}}
  \vspace{-1mm}
  \centerline{\small{(f) TMS-CSC}}
\end{minipage}
\hfill
\begin{minipage}{0.242\linewidth}
  \centerline{\includegraphics[width=1\linewidth]{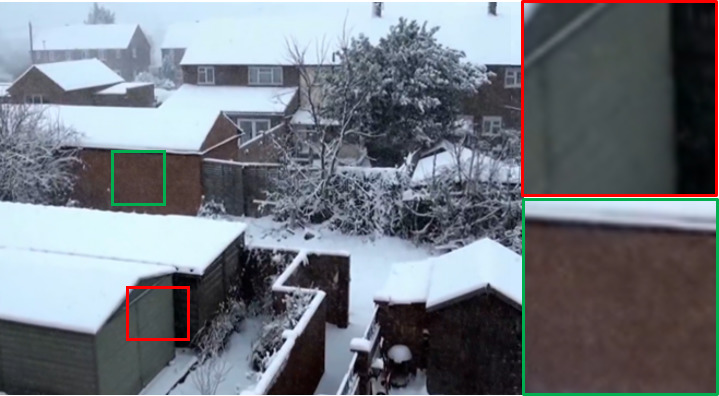}}
  \vspace{-1mm}
  \centerline{\small{(g) OTMS-CSC}}
\end{minipage}
\caption{
(a) Two input frame of a real snowy video with obvious illumination variation. (b)-(h) Recovered frames obtained by different competing methods.
}
\label{fig:tlondon}
\vspace{-2mm}
\end{figure*}

\begin{figure*}[!htb]\vspace{-2mm}
\begin{minipage}{0.242\linewidth}
  \centerline{\includegraphics[width=1\linewidth]{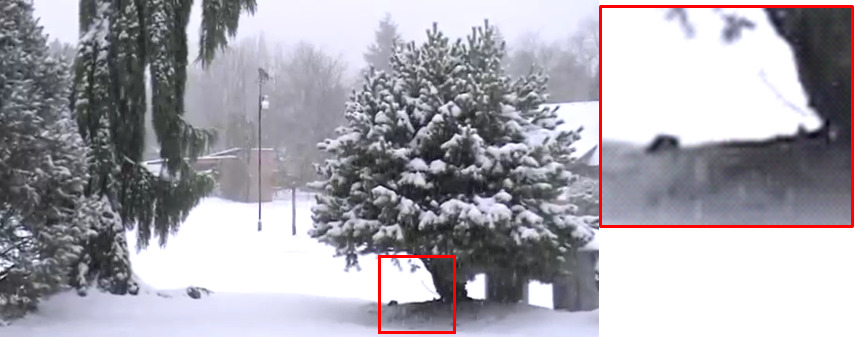}}
\end{minipage}
\hfill
\begin{minipage}{0.242\linewidth}
  \centerline{\includegraphics[width=1\linewidth]{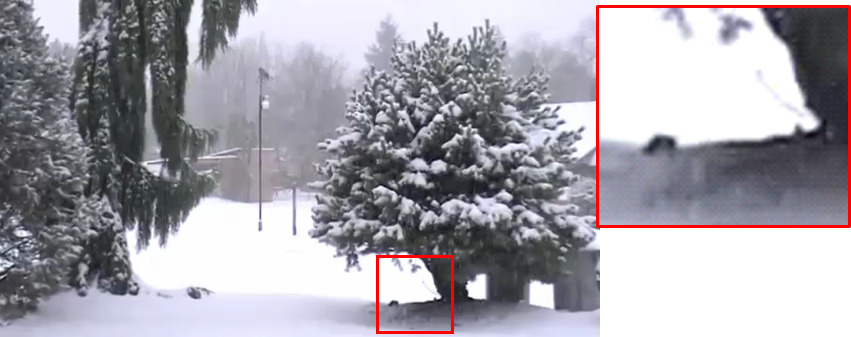}}
\end{minipage}
\hfill
\begin{minipage}{.242\linewidth}
  \centerline{\includegraphics[width=1\linewidth]{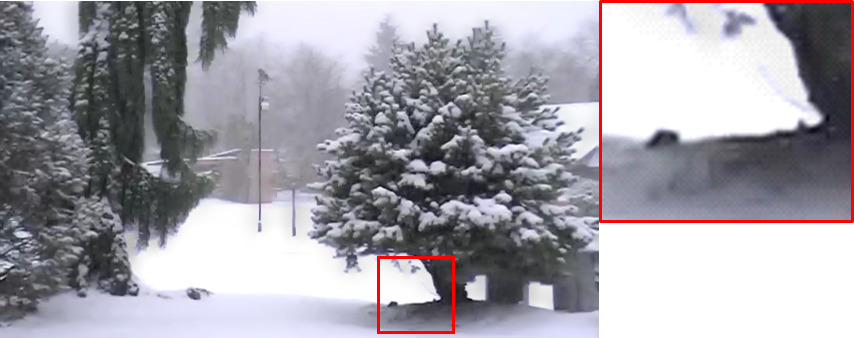}}
\end{minipage}
\hfill
\begin{minipage}{0.242\linewidth}
  \centerline{\includegraphics[width=1\linewidth]{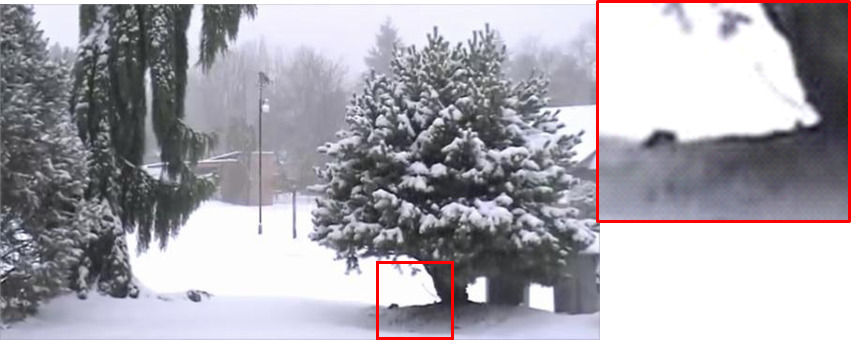}}
\end{minipage}
\vfill
\begin{minipage}{0.242\linewidth}
  \centerline{\includegraphics[width=1\linewidth]{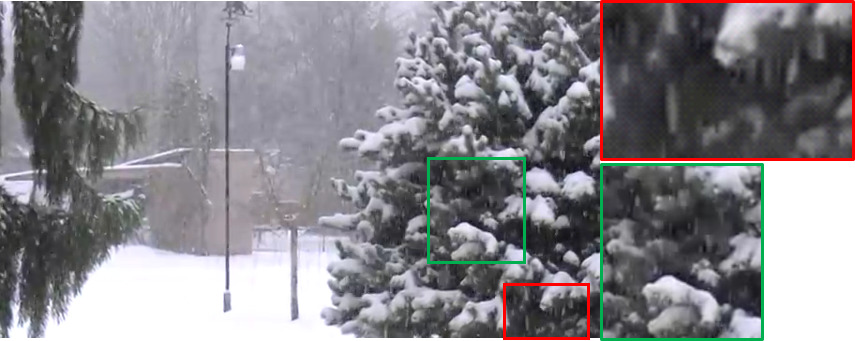}}
  \vspace{-1mm}
  \centerline{\small{(a) Input}}
\end{minipage}
\hfill
\begin{minipage}{0.242\linewidth}
  \centerline{\includegraphics[width=1\linewidth]{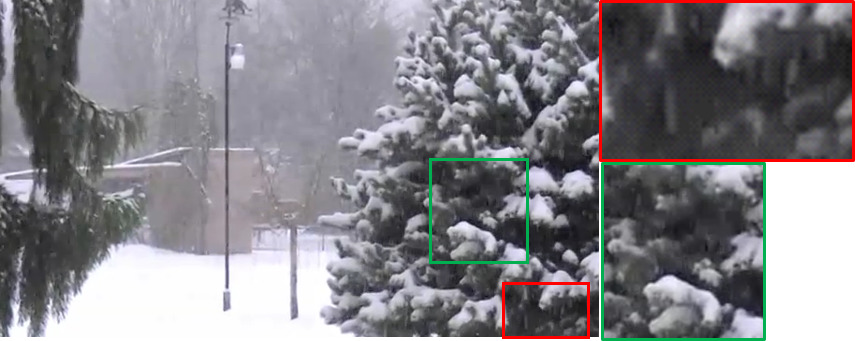}}
  \vspace{-1mm}
  \centerline{\small{(b) Garg et al. \cite{Garg2004Detection} }}
\end{minipage}
\hfill
\begin{minipage}{0.242\linewidth}
  \centerline{\includegraphics[width=1\linewidth]{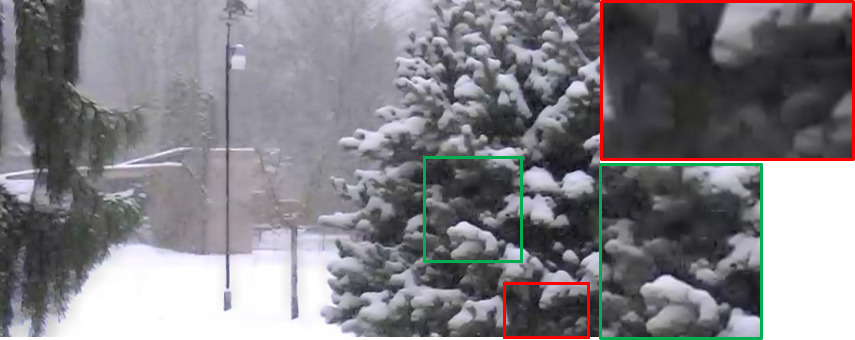}}
  \vspace{-1mm}
  \centerline{\small{(c) Jiang et al. \cite{jiang2017cvpr}}}
\end{minipage}
\hfill
\begin{minipage}{0.242\linewidth}
  \centerline{\includegraphics[width=1\linewidth]{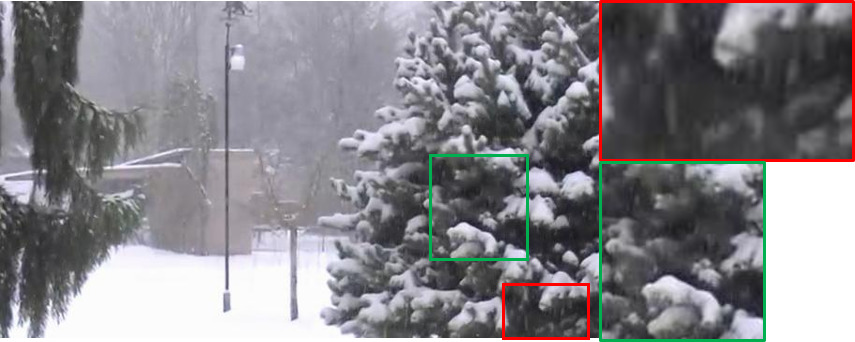}}
  \vspace{-1mm}
  \centerline{\small{(d) Ren et al. \cite{ren2017cvpr}}}
\end{minipage}
\vfill
\begin{minipage}{0.242\linewidth}
  \centerline{\includegraphics[width=1\linewidth]{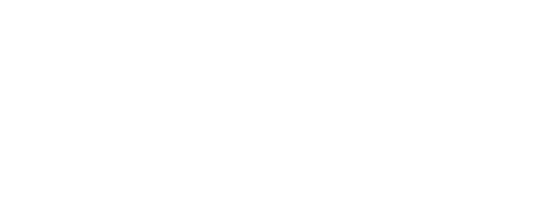}}
  \vspace{-1mm}
\end{minipage}
\hfill
\begin{minipage}{0.242\linewidth}
  \centerline{\includegraphics[width=1\linewidth]{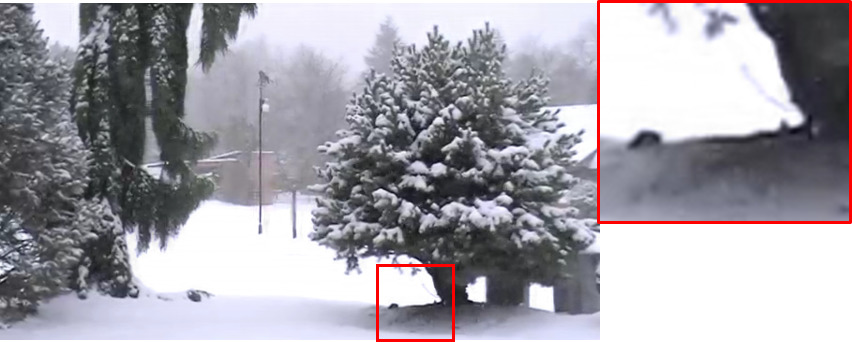}}
\end{minipage}
\hfill
\begin{minipage}{0.242\linewidth}
  \centerline{\includegraphics[width=1\linewidth]{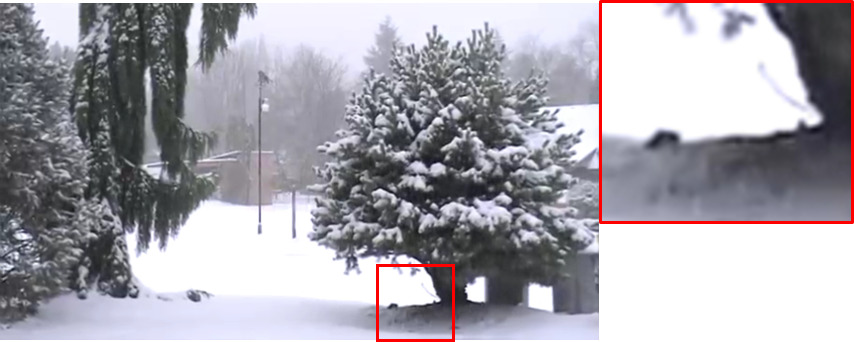}}
\end{minipage}
\hfill
\begin{minipage}{0.242\linewidth}
  \centerline{\includegraphics[width=1\linewidth]{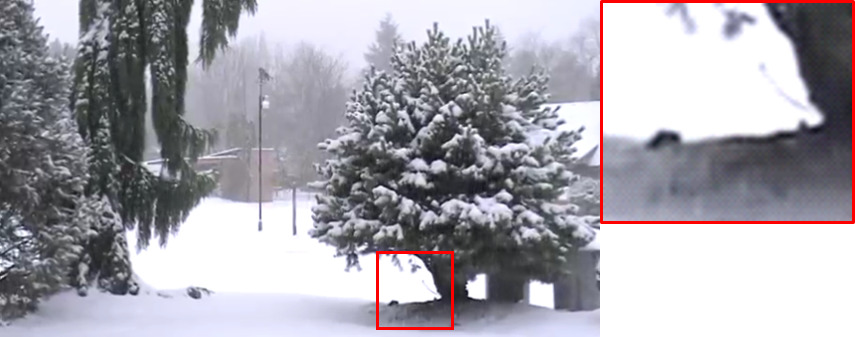}}
\end{minipage}
\vfill
\begin{minipage}{0.242\linewidth}
  \centerline{\includegraphics[width=1\linewidth]{s_pine//9}}
  \vspace{-1mm}
\end{minipage}
\hfill
\begin{minipage}{0.242\linewidth}
  \centerline{\includegraphics[width=1\linewidth]{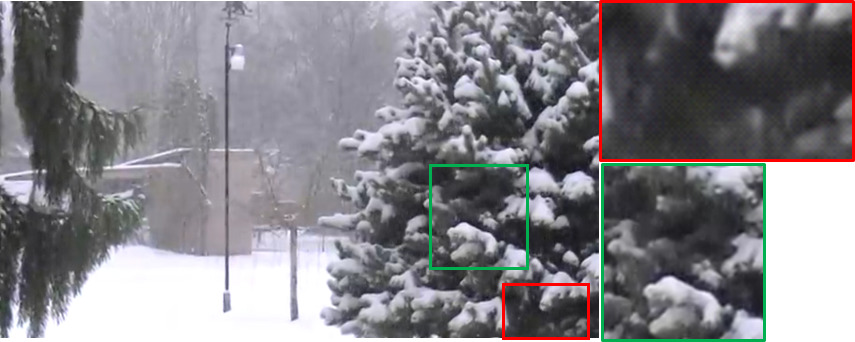}}
  \vspace{-1mm}
  \centerline{\small{(e) Liu et al. \cite{Liu18Erase}}}
\end{minipage}
\hfill
\begin{minipage}{0.242\linewidth}
  \centerline{\includegraphics[width=1\linewidth]{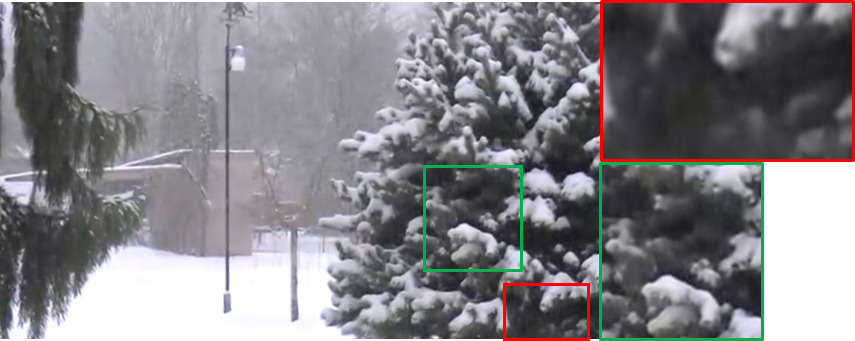}}
  \vspace{-1mm}
  \centerline{\small{(f) TMS-CSC}}
\end{minipage}
\hfill
\begin{minipage}{0.242\linewidth}
  \centerline{\includegraphics[width=1\linewidth]{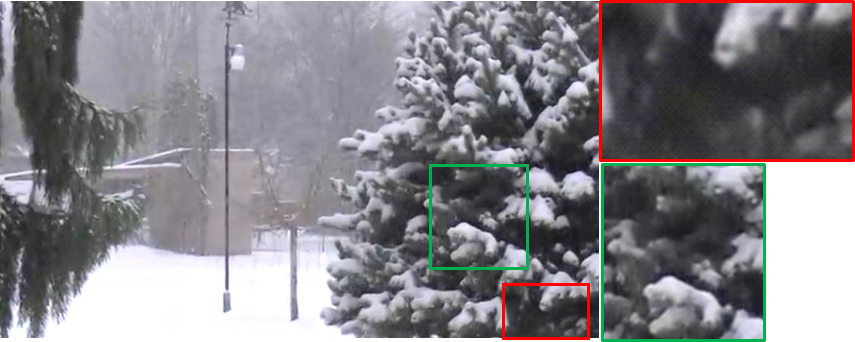}}
  \vspace{-1mm}
  \centerline{\small{(g) OTMS-CSC}}
\end{minipage}\vspace{-2mm}
\caption{
(a) Two input frame of a real snowy video with scale transformation. (b)-(g) Recovered frames obtained by different competing methods.}
\label{fig:spine}
\vspace{-3mm}
\end{figure*}

\begin{figure*}[!htb]\vspace{-1mm}
\begin{minipage}{0.242\linewidth}
  \centerline{\includegraphics[width=1\linewidth]{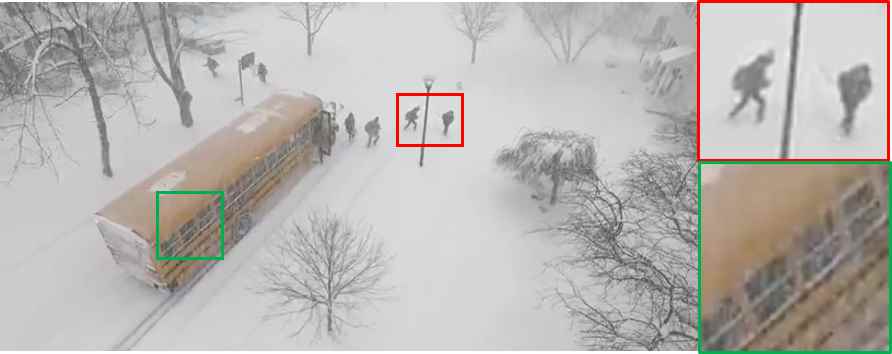}}
\end{minipage}
\hfill
\begin{minipage}{0.242\linewidth}
  \centerline{\includegraphics[width=1\linewidth]{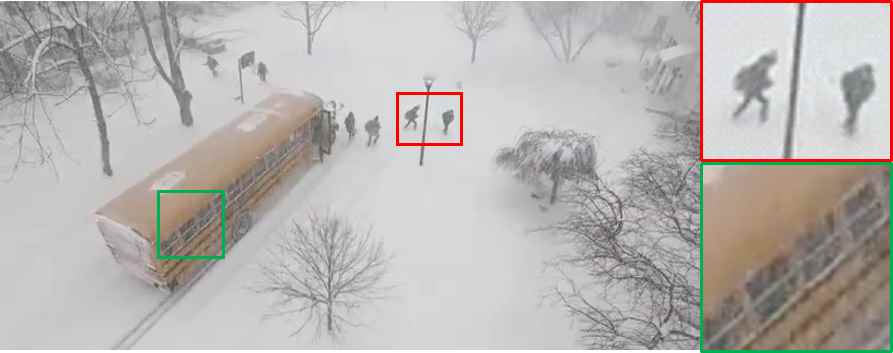}}
\end{minipage}
\hfill
\begin{minipage}{.242\linewidth}
  \centerline{\includegraphics[width=1\linewidth]{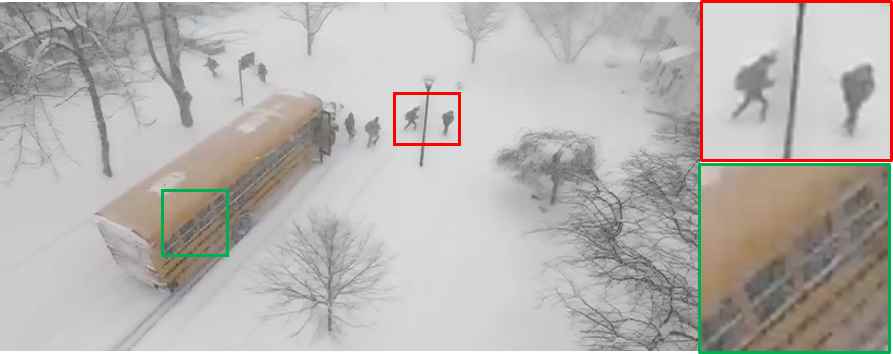}}
\end{minipage}
\hfill
\begin{minipage}{0.242\linewidth}
  \centerline{\includegraphics[width=1\linewidth]{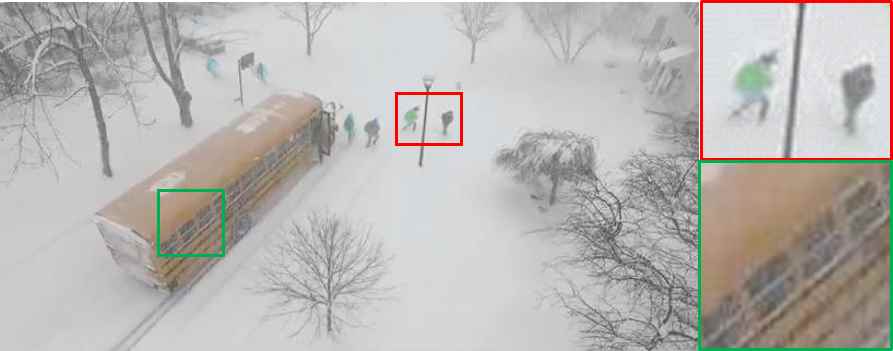}}
\end{minipage}
\vfill
\begin{minipage}{0.242\linewidth}
  \centerline{\includegraphics[width=1\linewidth]{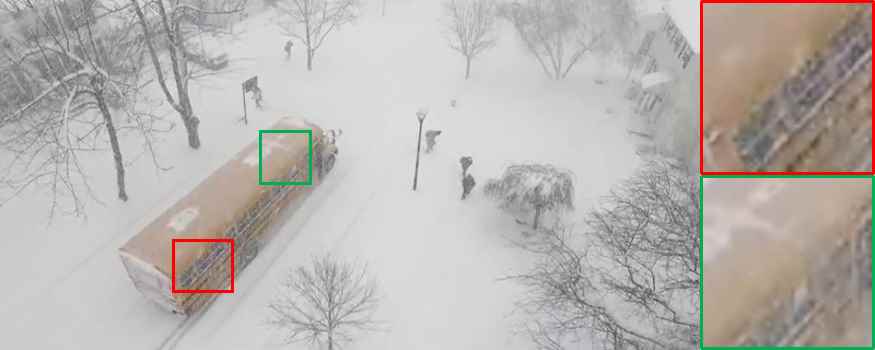}}
  \vspace{-1mm}
  \centerline{\small{(a) Input}}
\end{minipage}
\hfill
\begin{minipage}{0.242\linewidth}
  \centerline{\includegraphics[width=1\linewidth]{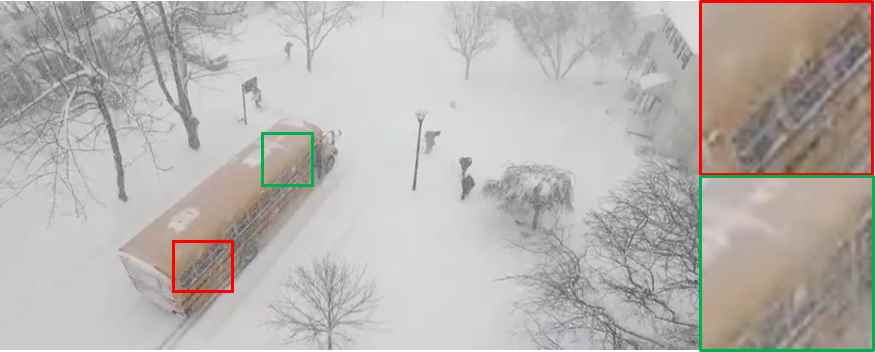}}
  \vspace{-1mm}
  \centerline{\small{(b) Garg et al. \cite{Garg2004Detection} }}
\end{minipage}
\hfill
\begin{minipage}{0.242\linewidth}
  \centerline{\includegraphics[width=1\linewidth]{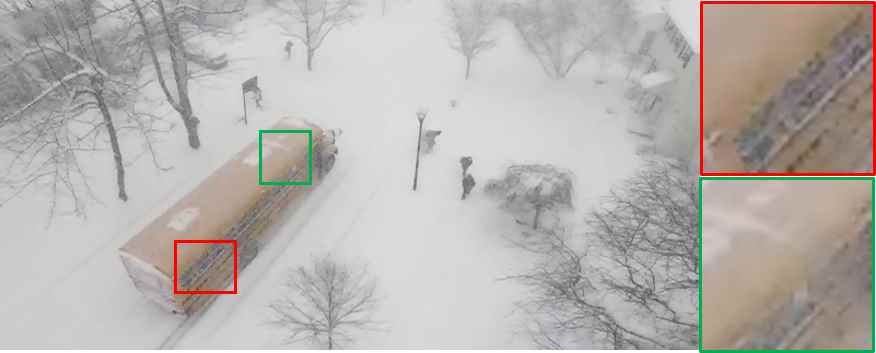}}
  \vspace{-1mm}
  \centerline{\small{(c) Jiang et al. \cite{jiang2017cvpr}}}
\end{minipage}
\hfill
\begin{minipage}{0.242\linewidth}
  \centerline{\includegraphics[width=1\linewidth]{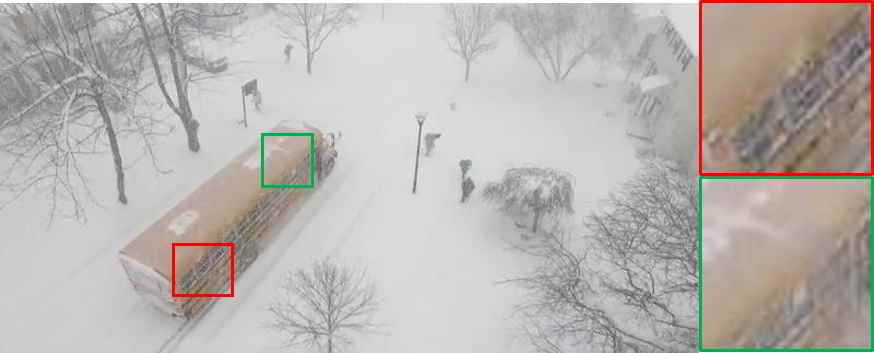}}
  \vspace{-1mm}
  \centerline{\small{(d) Ren et al. \cite{ren2017cvpr}}}
\end{minipage}
\vfill
\begin{minipage}{0.242\linewidth}
  \centerline{\includegraphics[width=1\linewidth]{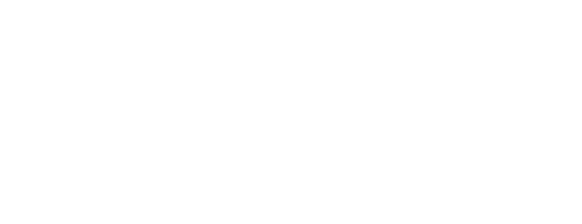}}
\end{minipage}
\begin{minipage}{0.242\linewidth}
  \centerline{\includegraphics[width=1\linewidth]{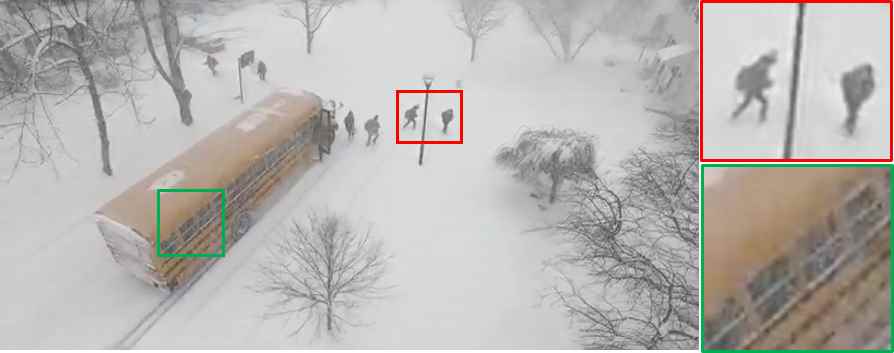}}
\end{minipage}
\hfill
\begin{minipage}{0.242\linewidth}
  \centerline{\includegraphics[width=1\linewidth]{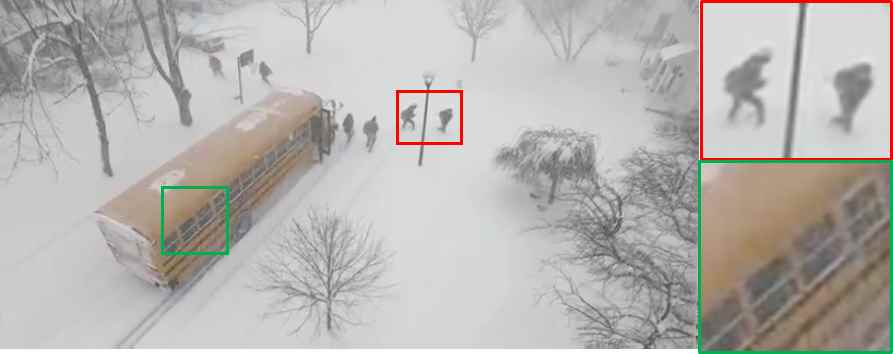}}
\end{minipage}
\hfill
\begin{minipage}{0.242\linewidth}
  \centerline{\includegraphics[width=1\linewidth]{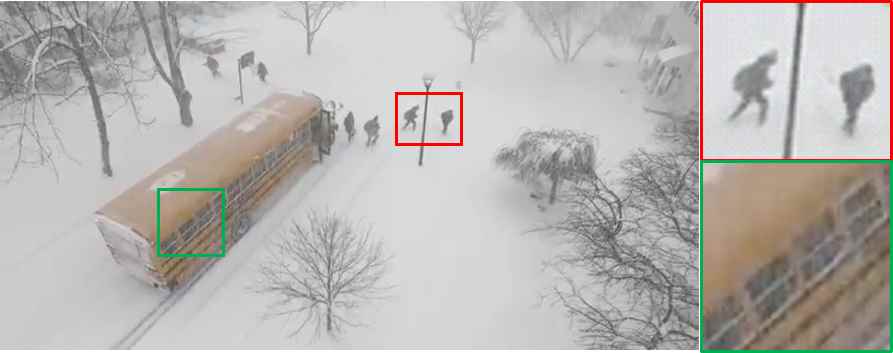}}
\end{minipage}
\vfill
\begin{minipage}{0.242\linewidth}
  \centerline{\includegraphics[width=1\linewidth]{aerial//9}}
\end{minipage}
\hfill
\begin{minipage}{0.242\linewidth}
  \centerline{\includegraphics[width=1\linewidth]{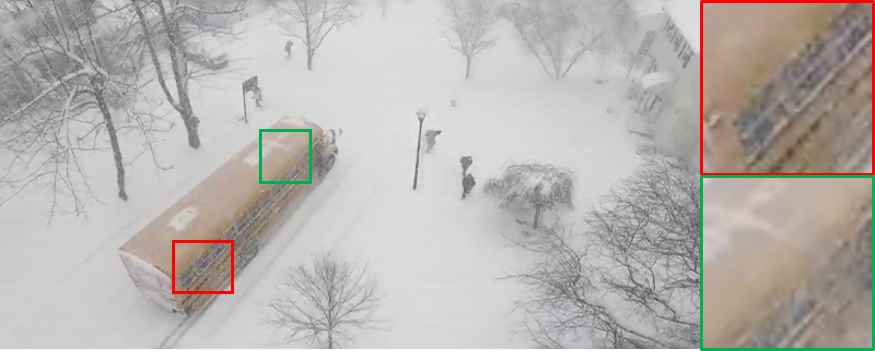}}
  \vspace{-1mm}
  \centerline{\small{(e) Liu et al. \cite{Liu18Erase}}}
\end{minipage}
\hfill
\begin{minipage}{0.242\linewidth}
  \centerline{\includegraphics[width=1\linewidth]{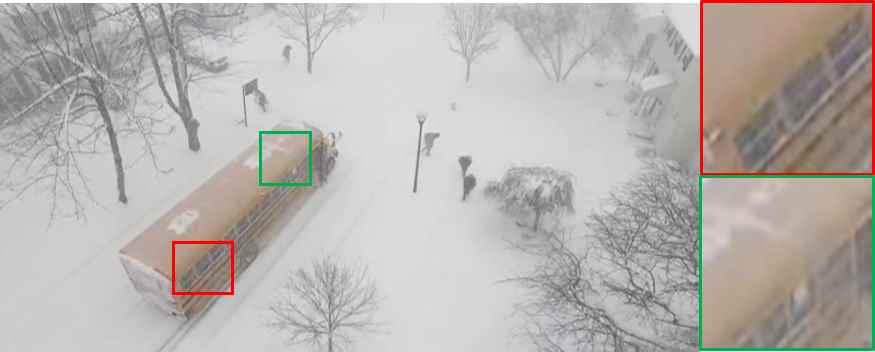}}
  \vspace{-1mm}
  \centerline{\small{(f) TMS-CSC}}
\end{minipage}
\hfill
\begin{minipage}{0.242\linewidth}
  \centerline{\includegraphics[width=1\linewidth]{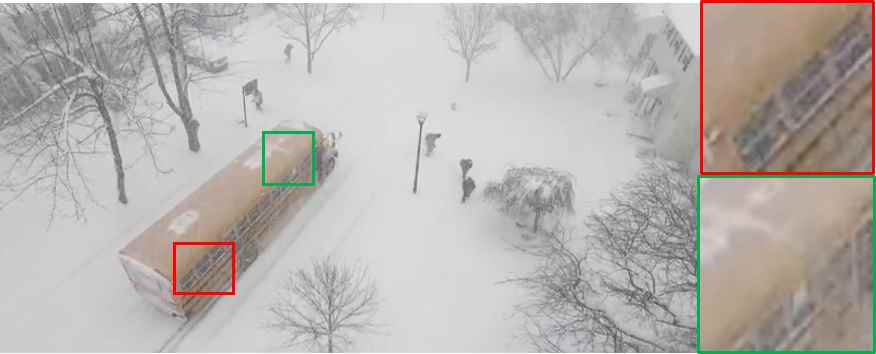}}
  \vspace{-1mm}
  \centerline{\small{(g) OTMS-CSC}}
\end{minipage}
\caption{
(a) Two input frame of a real aerial video with complex moving objects. (b)-(g) Recovered frames obtained by different competing methods.
}
\label{fig:aerial}
\vspace{-3mm}
\end{figure*}

\subsection{Experiments on Videos with Real Rain/Snow}
We further evaluate the performance of the proposed method on videos with real rainy or snowy scenarios. Eight real videos have been included in our experiments, including three captured under static backgrounds (as shown in Fig.~\ref{fig:compfinal}--\ref{fig:animal}) and five under dynamic backgrounds (as shown in Fig.~\ref{fig:light}--\ref{fig:aerial}) with typical transformations like random jitter, translation, scale transformation and aerial view. Fig.~\ref{fig:compfinal} and Fig.~\ref{fig:postbox} are two public rain videos both used in \cite{Garg2005When}, and the videos of Fig.~\ref{fig:night}, Fig.~\ref{fig:light}, Fig.~\ref{fig:tlondon} and Fig.~\ref{fig:spine} are downloaded from Youtube\footnote{https://www.youtube.com/watch?\{v=KzEv1h-JgaY, v=kNTYEKjXqzs, v=wb3gWRcKyCI, v=HbgoKKj7TNA\}} respectively.

The videos shown in Fig.~\ref{fig:compfinal} and Fig.~\ref{fig:night} are
captured by surveillance equipments in street, containing dynamically varying rain structures along time. From the figures, we can easily observe that
the derained frames of all other compared methods still contain certain rain streaks and the extracted rain layer is mixed with edges from the background. By contrast, the OTMS-CSC method, as well as MS-SCS, is capable of better removing all the rain streaks without mixing extra information into the rain layer.

Fig.~\ref{fig:animal} and Fig.~\ref{fig:light} show two real snowy video sequences captured on a real scene with poor visibility containing dynamic backgrounds. It is easy to see from the figures that most other competing methods have degenerated performance in snow removing, especially in the area around the light. Comparatively, our method can finely remove the snow and preserve the texture detail of the frame.

Fig.~\ref{fig:postbox} and Fig.~\ref{fig:tlondon} show the snow removal results on real videos with fast horizontal movement and obvious illumination variations, respectively. From Fig. \ref{fig:postbox}, it can be seen that the methods proposed by Garg et al. and Jiang et al. cannot fully remove the snow and recover the texture information underlying the frames. The methods proposed by Ren et al. and Liu et al. fail to detect and remove the snowflakes since they are not capable of dealing with video transformations. The OTMS-CSC method, as well as TMS-CSC, can obtain better visualized performance since they consider the background transformation in the modeling. This verifies that aligning the video background can help to improve the final performance of snow removal especially for dynamic videos.

Fig.~\ref{fig:spine} and Fig.~\ref{fig:aerial} show two challenging real snowy videos. Fig.~\ref{fig:spine} is captured in the condition of light snow and most backgrounds are covered with white snow, and thus it is not easy even for humans to observe the falling snow in a frame. Fig.~\ref{fig:aerial} is also challenging since it is a aerial video and with evident scale variations across frames. It is seen from the figures that our method can still perform relatively satisfactory in these videos, which verifies its robustness in real cases. Please refer to the website\textsuperscript{\ref{footnote1}} for more comprehensive illustration of the video results.

\subsection{Run time comparison}
\begin{figure}[!htb]
	\begin{minipage}{0.49\linewidth}
		\centerline{\includegraphics[width=1\linewidth]{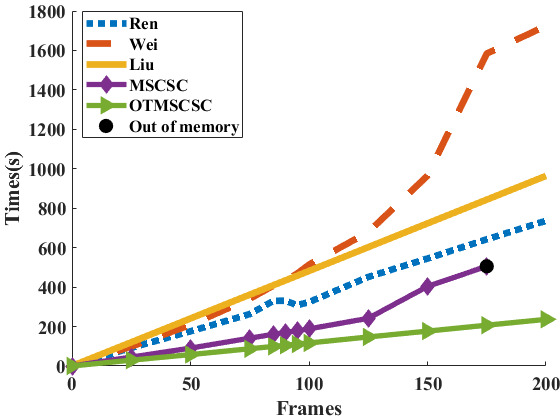}}
		\vspace{-1mm}
		\centerline{\footnotesize{(a) Fig. \ref{fig:4080} }}
	\end{minipage}
    \hfill
    \begin{minipage}{0.49\linewidth}
		\centerline{\includegraphics[width=1\linewidth]{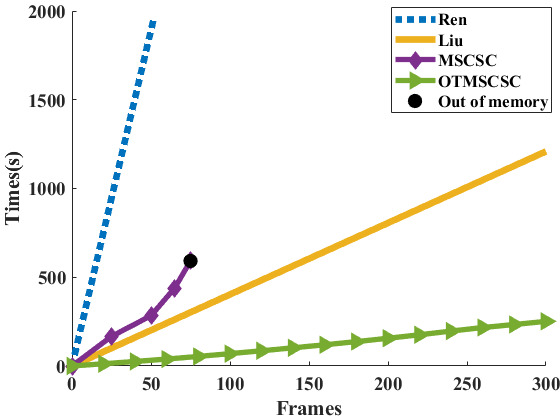}}
		\vspace{-1mm}
		\centerline{\footnotesize{(b) Fig. \ref{fig:human}}}
	\end{minipage}
	\vfill
	\begin{minipage}{.49\linewidth}
		\centerline{\includegraphics[width=1\linewidth]{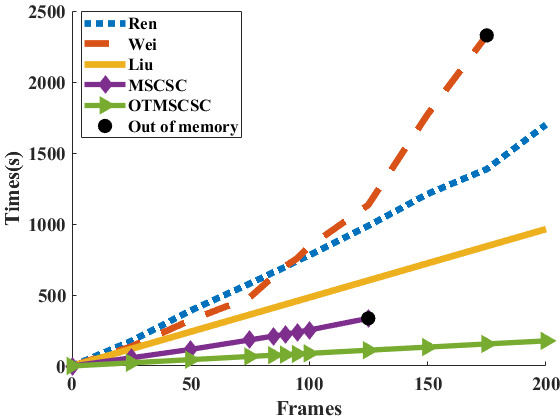}}
		\vspace{-1mm}
		\centerline{\footnotesize{(c) Fig. \ref{fig:light}}}
	\end{minipage}
	\hfill
	\begin{minipage}{.49\linewidth}
		\centerline{\includegraphics[width=1\linewidth]{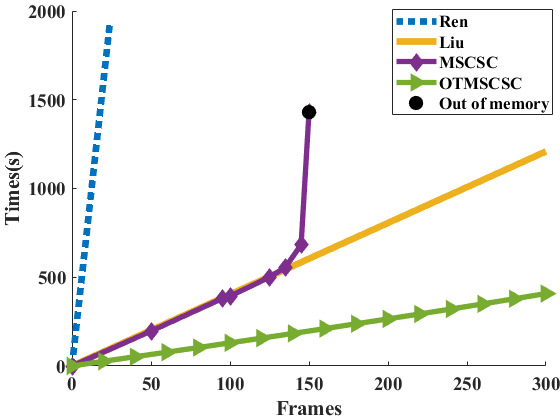}}
		\vspace{-1mm}
		\centerline{\footnotesize{(d) Fig. \ref{fig:spine}}}
	\end{minipage}
    \vspace{-2mm}
	\caption{ Run time comparison of comparable methods on several videos. The black point denotes the method over the current frames will report an out of memory error.}
	\vspace{-5mm}
	\label{fig:times}
\end{figure}

To show the efficiency of the proposed online method, we list the average running time per frame of each compared method in Table~\ref{timestable} in four representative static and dynamic videos with synthetic and real rain/snow, respectively. From the table, the speed advantage of the OTMS-CSC method is evident attributed to its online learning manner. Besides, as we show in Fig.~\ref{fig:times}, this online method has a good scalability, i.e., its time cost is linearly increasing with more input video frames, naturally due to its fixed training time on each video frame. Together with its fixed space complexity along time as discussed in Sec. 3.4.3, the method is expected to be potentially useful for real streaming videos.

\begin{table}[!htbp]
	\caption{  Run time comparison of all competing methods on four typical rainy/snowy videos.}\label{timestable}
	\vspace{-8mm}
	\begin{center}
		\setlength{\tabcolsep}{0.5mm}{
			\linespread{2}
			\begin{tabular}{c|cc|cccccp{0.5cm}}
				\Xhline{1.5pt}
				Type &Dataset &Size &Ren \cite{ren2017cvpr}&Wei \cite{wei2017should}&Liu \cite{Liu18Erase}&MS-CSC&OTMS-CSC\\
				\hline
				\multirow{2}{*}{Static} &Fig. \ref{fig:4080}    &{\scriptsize {$270\times480$}} 	&3.67	&8.62	&4.82	&3.37	&{\textbf{0.96}} \\ 
				&Fig. \ref{fig:light}   &{\scriptsize {$360\times480$}} 	&8.05	&13.30	&4.82	&2.69	&{\textbf{0.88}}	\\ 
				\hline
				\multirow{2}{*}{Dynamic}&Fig. \ref{fig:human}   &{\scriptsize {$288\times352$}} 	&50.3	&-       &4.03	&9.53	&{\textbf{0.87}}\\
				&Fig. \ref{fig:spine}   &{\scriptsize {$360\times640$}} 	&80.4  & - 	&8.55	&23.47	&{\textbf{1.36}} \\ 
				\Xhline{1.5pt}
			\end{tabular}}
		\end{center}
		\vspace{-5mm}
\end{table}

\section{Conclusion}
In this paper, we have proposed a new rain/snow removal method for surveillance videos containing dynamic rain/snow captured with camera jitter.
Both dynamic characteristics of rain/snow variations and background scenes along time inevitably encountered in real cases, have been fully considered in our method. Especially, the method is with a natural online implementation manner, with fixed space and time complexity for handling each frame of continuously coming videos, making it potentially useful for dealing with practical streaming video sequences. In the future, we will further ameliorate the capability of the proposed method in more challenging video cases, like those captured under moving cameras or those under background with strong color contrast and rain/snow with large streak shapes,
and try to design rational techniques or use some advanced computing equipments to further speed up the method for each unique frame to make it meet with the real-time requirements on practical streaming videos.



\ifCLASSOPTIONcaptionsoff
  \newpage
\fi

\bibliographystyle{IEEEtran}
\bibliography{OTMSCSC}

\end{document}